\documentclass[a4paper]{cas-dc}

\usepackage[numbers,sort]{natbib}

\usepackage{amssymb}
\usepackage{amsmath}
\usepackage{mathtools} 
\usepackage{textcomp}
\usepackage{tabularx}
\usepackage{makecell}
\usepackage{booktabs}
\usepackage{subfig}
\usepackage{relsize}
\usepackage{placeins}
\usepackage{seqsplit}
\usepackage{url}
\usepackage{hyperref}
\usepackage{xurl}
\usepackage{xcolor}
\usepackage{multirow}
\usepackage{float}
\usepackage{placeins}

\definecolor{prefillcol}{HTML}{55A868}
\definecolor{kvcol}{HTML}{E2C044}
\definecolor{nexttokcol}{HTML}{DD8452}

\newcommand{\termbox}[2]{\setlength{\fboxrule}{1.2pt}\setlength{\fboxsep}{3pt}%
  \fcolorbox{#1}{white}{$\displaystyle
    \vphantom{\left(\tau_{\text{in}}+\frac{\tau_{\text{out}}}{2}\right)}%
    #2$}}
  
\usepackage[most]{tcolorbox}
\definecolor{takeawaygreen}{HTML}{55A868}

\newtcolorbox{takeaway}[2]{
  float=tbp,
  floatplacement=tbp,
  enhanced,
  colback=takeawaygreen!8!white,
  colframe=takeawaygreen!70!black,
  colbacktitle=takeawaygreen!70!black,
  coltitle=white,
  fonttitle=\bfseries,
  title={RQ#1: \normalfont #2},
  boxrule=0.8pt,
  arc=2mm,
  left=2mm, right=2mm, top=1mm, bottom=1mm
}

\newif\ifdraft
\drafttrue

\ifdraft
  \newenvironment{revision}{\color{red}}{}
\else
  
\fi

\def\tsc#1{\csdef{#1}{\textsc{\lowercase{#1}}\xspace}}
\tsc{WGM}
\tsc{QE}
\tsc{EP}
\tsc{PMS}
\tsc{BEC}
\tsc{DE}

\begin{document}
\let\WriteBookmarks\relax
\def\floatpagepagefraction{1}
\def\textpagefraction{.001}

\shorttitle{The SIMPLE Law for Energy Cost of LLM Inference}

\shortauthors{H. Pizzini Cavagna, A. Proia et~al.}

\title [mode = title]{The SIMPLE Law for Energy Cost of LLM Inference}





%




\author[1]{Hiari {Pizzini Cavagna}}[
orcid=0009-0005-2768-0418
]
\cormark[1]
\fnmark[1]
\ead{hiari.pizzinicavagna@unibo.it}

\author[2]{Andrea Proia}[
]
\fnmark[1]

\author[3]{Daniele Cesarini}[
]

\author[2]{Zeynep Kiziltan}[
]

\author[1]{Andrea Bartolini}[
]

\fntext[1]{These authors contributed equally to this work.}




\affiliation[1]{organization={DEI - University of Bologna},
    city={Bologna},
    country={Italy}}

\affiliation[2]{organization={DISI - University of Bologna},
    city={Bologna},
    country={Italy}}
    
\affiliation[3]{organization={CINECA},
    city={Casalecchio di Reno},
    country={Italy}}
    
\cortext[cor1]{Corresponding author}



\begin{abstract}
Generative AI is billed per token, yet tokens are not equally expensive. 
Each token enters the KV-cache and raises the cost of those that follow, so energy grows non-linearly along a sequence whose tokens look identical.
No analytical model states what a token costs without per-LLM fitting, 
and no law names the factors that set it. 
From closed-form expressions of LLM inference's computational and memory-access complexity, we derive the
\textit{\textbf{S}IMPLE \textbf{I}nference \textbf{M}odel to \textbf{P}redict \textbf{L}LMs \textbf{E}nergy} (\textit{SIMPLE}), a four-parameter law that separates prefill from decode, and decode into KV-cache and next-token computation cost.
Over traces of  13 dense
LLMs from several families 
on NVIDIA H100 GPUs, it attains $R^2 = 0.969$ and $11.97\%$ MAPE, 
3.1-$3.5\times$ below the SOTA, 
without per-LLM calibration.
\textit{SIMPLE} explains as well as predicts: 
prefill energy grows quadratically with hidden dimension, so larger LLMs suffer most from long inputs; 
and higher GQA KV grouping reduces the per-token energy 
linearly relative to standard MHA. 
Increasing the reasoning token budget pushes output length beyond the analytically derived energy minimum, driving super-linear energy growth; 
accuracy rises but energy rises faster, so Intelligence per Joule peaks at an intermediate budget and falls by up to $3.14\times$ across a $4.0\times$ 
range. 
Moving to agentic serving raises energy per request by $13.0\times$,
as both prefill and KV-cache terms grow with input length; 
prefix caching removes most of the prefill cost ($-93\%$), 
but not KV-cache cost,
so even a cached agentic request costs $4.8\times$ more. 
\end{abstract}


\maketitle

\section{Introduction}
\label{sec:introduction}

Inference, not training, is where most of the energy of a Large Language Model (LLM) is spent: a model is trained once and then served continuously, and inference accounts for up to 90\% of its life-cycle energy~\cite{jegham2025hungryaibenchmarkingenergy}. Yet for an LLM not yet deployed, the only dependable way to know that cost is to build it, serve it, and measure it, an answer that arrives once the architecture is no longer a choice. The target is also moving: reasoning spends additional tokens to buy accuracy~\cite{s1-test-time}, while agentic workflows~\cite{copilot2025github} have pushed median prompt lengths from 750--1050 to 68K tokens~\cite{liu2026agentic}.

The token is the unit of account of this economy. Inference is priced, billed, budgeted, and benchmarked per token~\cite{mlperf, suscom_ai_emissions}, and capacity is planned in tokens per second and tokens per joule. Yet no analytical model states what a token costs, and no law names the factors that set it. A token is an observable of the workload, not a unit of computational work: each token enters the KV-cache and raises the cost of those that follow, so average token energy depends on input length, generation length, context reuse, and how attention is organized. The consequence is already visible in the economics: using a model with a lower listed price per token can incur up to $28\times$ the total cost of using a nominally more expensive one~\cite{reversal_stoica}. An entire industry settles in a currency whose production cost cannot be computed.

Energy models exhibit the same gap in analytical form. Prior work fits energy as a function of input and output lengths~\cite{vellaisamy, zhan, wilkins, oviedo} empirically, without deriving it from the computational kernels, thereby describing a measured regime rather than explaining it. \textit{SweetSpot}~\cite{sweetspot}, the most accurate analytical formulation reported to date,  removes the decisive obstacle by deriving the decode phase in closed form from the computation and memory accesses of decoder-only Transformer inference. \textit{SweetSpot}, however, is a fit, not a law. Its coefficients are identified on a single LLM and carry that LLM's architecture implicitly: the number of layers, the hidden dimension, and the organization of attention never appear as terms. The formulation therefore neither names the factors that set the cost of a token nor applies to a model it has not already measured, and later analytical models~\cite{vartziotis, delavande, niu_law_nvidia} relax this only in part.

This limitation is felt where decisions are now being made: in energy efficiency. Intelligence per Watt relates delivered capability to the power that produced it~\cite{saadfalcon2026intelligenceXwatt}, and serving is reported in tokens per joule~\cite{mlperf}, but both are measured on systems that already exist. They rank deployments after the fact and cannot say which architecture would deliver the same capability for less, how many reasoning tokens are worth spending~\cite{s1-test-time}, or at what context length serving an agentic request ceases to be affordable~\cite{liu2026agentic}. A measured efficiency locates no optimum.  
What is therefore needed is a law expressed in the architecture's own hyperparameters, one that explains which of them govern the energy of a token, predicts that energy for an LLM never benchmarked, and thereby turns efficiency from something observed into something derived.

Modern LLMs share the decoder-only Transformer~\cite{vaswani2017} and differ in a handful of hyperparameters and attention architectures: Multi-Head Attention (MHA) has largely given way to Grouped-Query Attention (GQA)~\cite{ainslie2023GQA} and Multi-head Latent Attention (MLA)~\cite{deepseek_v2}, each cutting memory traffic and KV-cache size, and with them the energy per generated token~\cite{sweetspot}. Furthermore, long generations, 68K-token contexts, 
and mandatory prefix caching push today's requests precisely into the regime where the KV-cache term dominates: no law of token cost can be blind to 
attention.

Together, these considerations motivate the following open research questions: 

\begin{itemize}
        
    \item \textbf{RQ1}: Is there an analytical model that can accurately predict inference energy across LLMs with different architectures, including the attention mechanism?
    
    \item \textbf{RQ2}: Can an LLM-agnostic model remain compact and accurate enough to be a practical alternative to LLM-specific models, while generalizing to unseen LLM architectures and workloads?
    
    \item \textbf{RQ3}: 
    How do LLM size and architectural hyperparameters influence inference energy consumption?

    \item \textbf{RQ4}: 
    How does the reasoning token budget affect inference energy and Intelligence per Joule?
    
    \item \textbf{RQ5}: How does the energy cost of a single query shift from non-agentic to agentic workloads?
\end{itemize}

\noindent
This work answers these with the following contributions:

\begin{itemize}

\item  We present a taxonomy of state-of-the-art LLM inference energy models, classified by functional form, grounding theory, and ability to predict inference energy across LLMs with different architectures, including the attention mechanism. We find that the most accurate model, \textit{SweetSpot}~\cite{sweetspot}, absorbs architectural information implicitly into its fitted coefficients, requiring per-LLM calibration and making it neither LLM-agnostic nor attention-aware. The SOTA therefore lacks an 
analytical model capable of accurately predicting the energy consumed by a generic request for an arbitrary LLM.

\item To close this gap, we analytically derive closed-form expressions for the computational and memory-access complexity of LLM inference, covering both MHA and GQA mechanisms 
Building on these, we propose the \textit{SIMPLE} law, 
which separately models \textit{prefill} and \textit{decode} energy costs, further decomposing the latter into the self-attention \textit{KV-cache} cost and the \textit{next-token} computation (i.e., the feed-forward network).  Validated against a family of analytically derived models spanning different fitting granularities and modeling approaches,
\textit{SIMPLE} achieves the best accuracy-compactness trade-off among the evaluated models, reducing the error of SOTA models by $3.1$--$3.5\times$. An interactive implementation of \textit{SIMPLE} is available at \href{https://huggingface.co/spaces/ecs-lab/SIMPLE}{ecs-lab/SIMPLE}. 
 
\item 
Applying the \textit{SIMPLE} law across LLMs of different sizes, families, and architectures, we show that hidden dimension $d$ is the most sensitive architectural hyperparameter, as prefill energy scales quadratically with $d$, while layer depth  contributes linearly. Attention architecture matters strongly: 
higher GQA KV grouping reduces per-token energy 
linearly relative to standard MHA, making it an impactful architectural choice for reducing inference energy. 

\item 
We use the \textit{SIMPLE} law to analytically derive the output length $\tau_{out}$ that minimizes per-token energy (the \textit{sweet spot}), showing that increasing the reasoning token budget pushes $\tau_{out}$ beyond this minimum, driving super-linear energy growth. Although accuracy grows with the reasoning budget, energy grows faster, so  Intelligence Per Joule peaks at an intermediate budget and falls by up to $3.14\times$ at the lowest and highest budgets across a $4.0\times$ range.

\item Using the \textit{SIMPLE} law, we analyze the energy cost of agentic AI workloads and show that energy per request increases by $13.0\times$ relative to non-agentic ones, driven by the growth of both the prefill and \textit{KV-cache} terms with input length. While prefix caching removes most of the prefill cost (-$93\%$), the KV-cache term, accounting for $86\%$ of the remaining energy, is unaffected, so even a cached agentic request still costs $4.8\times$ more than a non-agentic one. 

\end{itemize}

We evaluate our models on a large dataset \cite{sweetspot} of TensorRT-LLM \cite{tensorrtllm} traces collected from a set of widely used dense LLMs running on NVIDIA H100. The included models span model sizes and architectures, covering OPT \cite{opt_models}, LLaMA \cite{llama_models}, Gemma \cite{gemma_models}, Falcon \cite{falcon_models}, Qwen2 \cite{qwen2}, and Granite \cite{granite_models}.

The rest of the paper is organized as follows. After providing background in Section~\ref{sec:background}, we review related work in Section~\ref{sec:related_works}.
Then we derive the floating point and memory operations of LLM inference in Section~\ref{sec:model_formulation}, and building on this, we present in Section~\ref{sec:model_derivation} \textit{SIMPLE}, together with the family of complexity-derived models used throughout the evaluation. In 
Section~\ref{sec:experimental_study} we experimentally validate \textit{SIMPLE}, and discuss
in Section~\ref{sec:discussion} how model size, architectural hyperparameters, reasoning, and agentic workloads impact energy consumption. Finally, we conclude  in Section~\ref{sec:conclusions}.
\section{Background}
\label{sec:background}

This section introduces the background needed for the analysis in this work. We briefly review the fundamentals of LLM architectures, their computational characteristics, and the metrics used throughout the paper.

\paragraph{\textbf{LLM Architecture}}
LLMs are built on the Transformer architecture~\cite{vaswani2017}, which has become the standard for generative AI. A Transformer block consists of two main components, the Multi-Head Self-Attention Mechanism (MHA) and the Feed-Forward Network (FFN), each wrapped in residual connections, layer normalization, and dropout. In MHA, the hidden representation is projected into several subspaces, corresponding to different \textit{heads}. Each head learns to capture a different type of dependency within the sequence, and the head outputs are then combined. This improves the model's expressivity and its ability to represent diverse relationships, but it also increases the number of operations in proportion to the number of heads.  To reduce this overhead, Multi-Query Attention (MQA) was proposed in 2019~\cite{shazer2019MQA}. It keeps multiple query heads but shares a single set of Key-Value (KV) projections across all of them, which drastically reduces memory usage and KV-cache size during decoding and makes inference more efficient. More recently, Grouped-Query Attention (GQA) was introduced in 2023~\cite{ainslie2023GQA} as a generalization of MHA and MQA: keys and values are shared within groups of heads, striking a balance between the efficiency of MQA and the expressivity of MHA. Multi-Head Latent Attention (MLA), introduced with DeepSeek-V2~\cite{deepseek_v2}, provides a different mechanism to reduce the KV-cache footprint. Rather than sharing KV projections across heads, it compresses keys and values into a single low-rank latent vector. The full per-head keys and values are then recovered from this latent through up-projection matrices, preserving the expressivity of MHA at a fraction of the memory cost. Despite its distinct compression mechanism, MLA can nonetheless be seen as a variant of GQA~\cite{gqa_vs_mla}.

\paragraph{\textbf{LLM Inference Computation}}
In large-scale models, most of the learnable weights, referred to as parameters, reside in the FFN layers, whereas the runtime complexity is dominated by the attention mechanism, owing to its quadratic dependence on the input sequence length. The total number of parameters, typically in the billions, is referred to as model size. 
LLM inference is typically split into two phases: \textit{prefill}, in which the entire input sequence is processed in parallel to initialize the KV-cache, and \textit{decode}, in which tokens are generated autoregressively \cite{llminfrence_unveiled}.
During prefill, the computational cost is dominated by the quadratic attention term; during decode, thanks to the reuse of cached keys and values, the cost per token grows only linearly with the sequence length. Nevertheless, for long outputs, the decode phase often accounts for most of the total inference time, since tokens must be produced one at a time, which restricts the opportunities for parallelization.
On modern GPUs, the throughput of LLM inference is constrained not only by arithmetic performance but also by memory bandwidth. Storing and retrieving the large KV-cache during decoding places considerable stress on the memory subsystem.
For example, in autoregressive decoding, every generated token requires loading all previous keys and values from memory, which makes attention kernels strongly memory-bound, especially for long contexts.

\paragraph{\textbf{Metrics and Measurement }}
We define the total energy consumption $E_{\text{tot}}(\tau_{\text{in}}, \tau_{\text{out}})$, expressed in Joules, as the energy the GPU consumes to process a given request, characterized by an input length $\tau_{\text{in}}$ and an output length $\tau_{\text{out}}$.
Normalizing $E_{\text{tot}}$ by $\tau_{\text{out}}$ yields the Energy-per-Token $E_{\text{tok}}$:
{
\[
E_{\text{tok}} = \frac{E_{\text{tot}}}{\tau_{\text{out}}} \quad \left[\frac{\mathrm{\text{Joules}}}{\mathrm{\text{Token}}}\right]
\]
}

We also define energy efficiency $E_{\text{eff}}$ as the reciprocal of $E_{\text{tok}}$, capturing the number of output tokens generated per unit of energy. 
Finally, to relate energy consumption to the quality of the generated output, we adopt the Intelligence per Joule (IPJ) metric \cite{saadfalcon2026intelligenceXwatt}. We compute it as the ratio of task accuracy to the energy required to process a request:
{
\[
\text{IPJ} = \frac{\text{Acc}}{E_{\text{tot}}} \quad \left[\frac{\mathrm{\text{Accuracy $\%$}}}{\mathrm{\text{Joule}}}\right]
\]
}
where $\text{Acc}$ is the accuracy averaged over the evaluated queries and $E_{\text{tot}}$ is the average per-request energy, making it a practical metric for realistic deployment scenarios. 

\section{Related Work}
\label{sec:related_works}
\begin{table*}[t]
\centering
\caption{Comparison of existing energy analytical and empirical models for LLM inference.}
\label{tab:related_models}

\renewcommand{\arraystretch}{1.3}
\begin{tabular}{llccc}
\toprule
\textbf{Model}
& \textbf{$E_{\text{tot}}$ Form}
& \textbf{Analytical}
& \textbf{LLM-agnostic}
& \textbf{\makecell[c]{Attention-aware}} \\
\midrule
Vellaisamy et al., 2026 ~\cite{vellaisamy}
    & Linear    & $\times$ & $\times$ & $\times$ \\
Zhan et al., 2026 ~\cite{zhan}
    & Linear    & $\times$ & $\times/\checkmark$ & $\times$ \\
Delavande et al., 2026 ~\cite{delavande}
    & Bilinear  & $\checkmark$ & $\times$  & $\times$ \\
Wilkins et al., 2025 ~\cite{wilkins}
    & Bilinear  & $\times$ & $\times$ & $\times$ \\
Oviedo et al., 2026 ~\cite{oviedo}
    & Power-law & $\times$ & $\times$ & $\times$ \\
SweetSpot, 2026 ~\cite{sweetspot}
    & Quadratic & $\checkmark$ & $\times$  & $\times$ \\
Vartziotis et al., 2026 ~\cite{vartziotis}
    & Quadratic & $\checkmark$  & $\checkmark$ & $\times$ \\
Niu et al., 2026 ~\cite{niu_law_nvidia}
    & \makecell[c]{-}        & $\checkmark$ & $\times$ & $\checkmark$ \\
\midrule
\textbf{SIMPLE}
    & Quadratic & $\checkmark$ &  $\checkmark$ & $\checkmark$ \\
\bottomrule
\end{tabular}
    
\end{table*}

Recent work on LLM inference energy falls broadly into two categories: (i) experimental characterization of energy consumption, and (ii) predictive models based on machine learning, analytical derivation, or empirical fitting. We review each in turn, then position our contributions with respect to the identified limitations. 

\subsection{Experimental Characterizations}
Several studies have characterized AI inference energy across different workloads and hardware conditions. Desislavov et al. \cite{suscom_trends_ai} characterize the relationship between  parameter count and energy consumption in deep neural networks.
More recently, 
Samsi et al. \cite{samsi2023wordswatts} benchmark energy for LLaMA models of varying sizes across two NVIDIA GPU generations (V100 and A100), including multi-node and multi-GPU execution. 
Chung et al. \cite{chung2025mlenergy} introduce \textit{ML.ENERGY}, a benchmark suite for measuring energy under realistic serving conditions across different tasks, and show that automated configuration recommendations can reduce consumption.
Stojkovic et al. \cite{stojkovic2024greenerllmsbringingenergyefficiency} study how inference server configurations affect latency, throughput, and energy. 
Maliakel et al. \cite{maliakel} examine the effect of GPU Dynamic Voltage and Frequency Scaling (DVFS) settings on energy for different workloads.
Fernandez et al. \cite{fernandez2025energyconsiderationslargelanguage} analyze energy across inference optimization strategies and task types. 
Wilhelm et al. \cite{wilhelm2025beyondtestime} investigate the accuracy–energy trade-off, focusing on how inference strategies such as chain-of-thought affect it. Ali et al. \cite{energy_accuracy_analysis} evaluate LLM energy consumption and accuracy across different tasks. 

Unlike the following works, none of these proposes an energy prediction model.

\subsection{Predictive Models}
Recent studies have proposed different approaches to predict the energy consumption of LLM inference. 
Machine-learning approaches learn this relationship from measurements: Castaño et al.~\cite{suscom_hardware_ai} train task-specific regressors to predict throughput and energy for Pareto-optimal hardware selection, while Aguirre-Usandizaga et al. \cite{suscom_LLM_forecasting} forecast short-term GPU power from HPC telemetry using time-series models. 
Fu et al.~\cite{llmco2} propose LLMCO2, a graph neural network that predicts the energy and operational carbon footprint of LLM inference from architectural and hardware-specific features.
Caravaca et al.~\cite{caravaca2025promptspowermeasuringenergy} develop regression-based predictors of prompt-level energy from measured workload, architectural, and hardware features.

These approaches trade interpretability for predictive flexibility. A complementary line of work instead expresses $E_{\text{tot}}$ as an explicit closed-form analytical or empirical formula.
These formulations focus on the number of input and output tokens ($\tau_{\text{in}}$ and $\tau_{\text{out}}$), while not explicitly accounting for batch size; which is commonly incorporated into predictive models targeting distributed inference \cite{maveriq, energylens}, whereas it is generally not considered in models for single-GPU inference, as in our setting.
The following paragraphs review the state-of-the-art models, grouped by $E_{\text{tot}}$ functional form and further characterized by three criteria that directly reflect the limitations our work addresses. 
First, whether they are analytically derived from first principles, such as Transformer computational complexity, or empirically. 
Second, whether they must be fit separately for each LLM. We refer to models that require a per-LLM fit as \emph{LLM-specific}, and to those that do not as \emph{LLM-agnostic}. 
Third, whether they are \emph{attention-aware}, meaning that they explicitly account for different attention mechanisms, rather than assuming standard MHA. Table~\ref{tab:related_models} summarizes all reviewed models along these criteria.

\paragraph{\textbf{Linear Models}}

A common empirical approach is to assign a fixed cost to input and output sequence lengths, yielding a linear model:
\begin{equation}
E_{\text{tot}} \approx \theta_0 + \theta_1\tau_{\text{in}} + \theta_2\tau_{\text{out}} \notag
\end{equation}
Both Vellaisamy et al.~\cite{vellaisamy} and Zhan et al.~\cite{zhan} propose linear models, though they differ in how they handle input-length dependence. 
Vellaisamy et al. focuses solely on decode energy absorbing the prefill cost into a fixed constant, effectively modeling decode energy as linear in $\tau_{\text{out}}$ alone, while Zhan et al. make $\tau_{\text{in}}$ an explicit term as in the formulation above. Neither model is derived from first principles or includes architectural hyperparameters, so both must be fitted separately for each LLM and are not attention-aware.

Zhan et al.~\cite{zhan} apply their model to Vision-Language Models and additionally propose an LLM-agnostic formulation that scales with model size $S$ (in billions of parameters):
\begin{equation}
E_{\text{tot}} \approx \theta_{0} + \theta_{1}S + \theta_{2}\tau_{\text{in}} + \theta_{3}\tau_{\text{out}} + \theta_{4}S\tau_{\text{in}} + \theta_{5}S\tau_{\text{out}} \notag
\end{equation}
While this formulation is LLM-agnostic, model size serves as the only architectural proxy, capturing scale but not structural hyperparameters such as the number of layers, hidden dimension, or attention configuration.

\paragraph{\textbf{Bilinear Models}}

Wilkins et al.~\cite{wilkins} and Delavande et al.~\cite{delavande} extend the linear formulation with an interaction term between input and output sequence lengths:
\begin{equation}
    E_{\text{tot}} \approx \theta_0 + \theta_1 \,\tau_{\text{in}} + \theta_2 \,\tau_{\text{out}}
    + \theta_3\,\tau_{\text{in}}\tau_{\text{out}} \notag
\end{equation}
Wilkins et al. motivate the interaction term statistically, using ANOVA to confirm that input and output lengths have a significant joint effect on energy, and fit the model empirically via ordinary least squares. Delavande et al. derive the same form analytically from the roofline model: the interaction term reflects the cumulative cost of attending to a growing context at each decode step. 
Neither model includes architectural hyperparameters, thus both must be fitted separately for each LLM and are not attention-aware.

\paragraph{\textbf{Power-law Models}}

Oviedo et al.~\cite{oviedo} propose a bottom-up framework that estimates per-query inference energy from node power draw and token throughput:
\begin{equation}
    E_{\text{tot}} = \frac{\text{PUE}}{3.6} \cdot
    \frac{P_{\text{node}} \cdot L_{\text{eff}}}{\text{TPS}(\tau_{\text{in}},\,\tau_{\text{out}})}
\notag
\end{equation}
where $\text{PUE}$ accounts for data-center overhead, $P_{\text{node}}$ is the measured steady-state node power, $\text{TPS}(\tau_{\text{in}}, \tau_{\text{out}})$ is fitted empirically to per-model benchmarks as a piecewise log-linear regression, and $L_{\text{eff}} \approx \tau_{\text{out}}$ scales energy by output length. Because $\text{TPS}$ is linear in $\log\tau_{\text{in}}$ and $\log\tau_{\text{out}}$, substituting it into the formula yields a power-law dependence on sequence lengths:
\begin{equation}
    E_{\text{tot}} \approx \theta_0\,\tau_{\text{in}}^{\theta_1}\,\tau_{\text{out}}^{\theta_2}
\notag
\end{equation}
The framework includes no architectural hyperparameters; thus the throughput model must be re-fitted for each LLM, making the approach LLM-specific and not attention-aware.

\paragraph{\textbf{Quadratic Models}}

\textit{SweetSpot}~\cite{sweetspot} models inference energy as proportional to FLOPs and data transfers, analytically derived from the transformer computational kernels as a function of $\tau_{\text{in}}$ and $\tau_{\text{out}}$. The decoding phase is handled by expressing the growing KV-cache context as an arithmetic series, which yields a closed-form quadratic formulation:
\begin{equation}
E_{\text{tot}} \approx \theta_{0} + \theta_{1}\,\tau_{\text{in}}
+ \theta_{2}\,\tau_{\text{in}}^{2}
+ \theta_{3}\,\tau_{\text{in}}\tau_{\text{out}}
+ \theta_{4}\,\tau_{\text{out}}
+ \theta_{5}\,\tau_{\text{out}}^{2} \notag
\end{equation}

The model reduces the MAPE of the bilinear baseline from over $30\%$ to below $2\%$. 
However, the fitted coefficients are LLM-specific, requiring an exhaustive measurement dataset for each target model. 
The approach therefore cannot generalize to unseen architectures or be applied during training. 
Moreover, the analytical derivation treats all $Q$, $K$, and $V$ projection matrices as having the same hidden dimension $d$, implicitly assuming standard MHA. Since the attention mechanism is not parameterized in the formulation, \textit{SweetSpot} is not attention-aware.

Following a similar analytical approach, Vartziotis et al.~\cite{vartziotis} decompose inference energy into compute and memory-access components:
\begin{equation}
E_{\text{tot}} \approx \alpha_{\text{TC}}\,C_{\text{TC}}
+ e_{\text{HBM}}\,B\,\eta(N) \notag
\end{equation}
where $C_{\text{TC}}$ denotes the total tensor-core FLOP count for prefill and decode, and $B$ denotes the total HBM traffic. $C_{\text{TC}}$ and $B$ are derived analytically from the transformer computational graph, following a similar decomposition to~\cite{sweetspot},  though the FLOP count $C_{\text{TC}}$ relies on the Kaplan training parameter-scaling approximation~\cite{kaplan} ($C \approx KNT$, $K = 6$ FLOPs per parameter per token) rather than an inference kernel-by-kernel tally. 
The factor $\eta(N)$, where $N$ is the number of model hyperparameters,  is a dimensionless inefficiency multiplier that accounts for non-ideal HBM bandwidth utilization, while $\alpha_{\text{TC}}$ and $e_{\text{HBM}}$ are hardware-specific energy coefficients for tensor-core and HBM operations, expressed in pJ/FLOP and pJ/bit respectively (H100 GPU, FP16/BF16). 
Because $\alpha_{\text{TC}}$ and $e_{\text{HBM}}$ are hardware constants rather than fitted per-model coefficients, and $C_{\text{TC}}$ and $B$ are expressed in terms of $N$ and structural hyperparameters such as layer count and hidden dimension, the model is LLM-agnostic and captures the order-of-magnitude energy and scaling trends observed in authors' measurements. 

However, its accuracy is limited, with a reported mean relative error of approximately $20\%$. 
Furthermore, expanding $C_{\text{TC}}$ and $B$ yields a long closed-form expression in the architectural hyperparameters, which makes the model hard to interpret and limits its usefulness for identifying the hyperparameters that drive energy consumption. Similarly to \textit{SweetSpot}, the model assumes MHA, so the reduced KV-cache traffic of GQA, which is standard in many modern LLMs, is not captured by the model.

To the best of our knowledge, Niu~et~al.~\cite{niu_law_nvidia} is the only existing work to propose an attention-aware energy model for LLM inference. Their model estimates the energy per output token of a single decode step,  expressing decode energy as a linear function of the bytes accessed, considering both the model weights and the KV-cache, and the measured decode time. The KV-cache byte term is analytically derived from the model architecture and attention mechanism: it is explicitly parametrized by the number of KV heads per layer ($N_{kv}$), which accounts for MHA and  GQA. The decode time, however,  must be measured at runtime. The resulting model requires a separate calibration and fit for each LLM and hardware configuration. Since the model covers only the decode phase energy, excluding the prefill, it is not directly applicable to predict the total inference energy $E_{\text{tot}}$ for a given query, which is the focus of this work. We therefore do not consider it further here.
\begin{takeaway}{1}{Is there an analytical model that can accurately predict inference energy across LLMs with different architectures, including the attention mechanism?}
No. The most accurate model, SweetSpot~\cite{sweetspot}, is LLM-specific: its coefficients must be fitted for each target LLM, so it cannot generalize to unseen architectures. LLM-agnostic alternatives do not explicitly model the attention mechanism; the most analytically grounded of these~\cite{vartziotis} reports a mean relative error of about 20\%  and yields hard-to-interpret expressions. The one model that parameterizes the attention mechanism~\cite{niu_law_nvidia} targets the decode phase only and requires measured runtime inputs, making it inapplicable to total inference energy prediction. No prior model is simultaneously interpretable, LLM-agnostic, attention-aware, and applicable to the full inference pipeline. \textit{SIMPLE} is the first to satisfy all these conditions.
\end{takeaway}
\subsection{SIMPLE Model}

The box above summarizes the answer to RQ1.  We address the key limitations of previous approaches by extending the derivation of the computational and memory-access complexities of \textit{SweetSpot}~\cite{sweetspot} to GQA, and using the resulting expressions to derive \textit{SIMPLE}, a compact LLM-agnostic model that explicitly represents the attention mechanism, builds on \textit{SweetSpot}’s interpretable quadratic structure to expose the role of individual architectural hyperparameters, and predicts energy across multiple LLM architectures without per-LLM calibration. 

Throughout our study, we target the Energy-per-Token $E_{\text{tok}}$, rather than total energy consumption $E_{\text{tot}}$, since normalizing by $\tau_{\text{out}}$ isolates the cost of generation from the cost of processing the input, exposing the trade-off between amortizing the prefill cost over more output tokens and incurring additional decode cost as $\tau_{\text{out}}$ grows. 
Moreover, $E_{\text{tok}}$ serves as a natural unit for comparison across different workloads and deployments, as it corresponds directly to operational metrics such as cost-per-token and tokens-per-joule, matching how inference is commonly priced and provisioned in production.

\section{FLOPs and MemOPs Formulation}
\label{sec:model_formulation}
\begin{table*}[ht]
\caption{Decoder-only LLM inference symbols and tensor dimensions.}
\centering
\renewcommand{\arraystretch}{1.15}
\small
\begin{tabular}{l l p{4cm} @{\hspace{1cm}} l l p{4cm}}
\toprule
\textbf{Symbol} & \textbf{Dimension} & \textbf{Description}
&
\textbf{Symbol} & \textbf{Dimension} & \textbf{Description} \\
\midrule

\multicolumn{3}{l}{\textbf{Scalars}}
&
\multicolumn{3}{@{}l}{\textbf{Prefill Phase}} \\
$\tau_{\text{in}}$  & $\mathbb{N}^{1\times1}$ & Number of input tokens
& $\mathbf{X}_P$    & $\mathbb{R}^{\tau_{\text{in}}\times d}$ & Input \\
$\tau_{\text{out}}$ & $\mathbb{N}^{1\times1}$ & Number of output tokens
& $\mathbf{Q}_P,\mathbf{O}_P$ & $\mathbb{R}^{\tau_{\text{in}}\times d}$ & Query, Output projections \\
$n_{\text{q}}$ & $\mathbb{N}^{1\times1}$ & Number of query heads
& $\mathbf{K}_P,\mathbf{V}_P$ & $\mathbb{R}^{\tau_{\text{in}}\times d_{\text{kv}}}$ & Key, Value projections \\
$n_{\text{kv}}$ & $\mathbb{N}^{1\times1}$ & Number of KV heads
& $\mathbf{S}_P$ & $\mathbb{R}^{\tau_{\text{in}}\times \tau_{\text{in}}}$ & Attention scores \\
$L$ & $\mathbb{N}^{1\times1}$ & Number of layers
& $\mathbf{A}_P$ & $\mathbb{R}^{\tau_{\text{in}}\times d}$ & Attention output \\
$d$ & $\mathbb{N}^{1\times1}$ & Hidden dimension
& $\mathbf{H}_P$ & $\mathbb{R}^{\tau_{\text{in}}\times4d}$ & Expanded activation \\
$R=\frac{n_{\text{kv}}}{n_{\text{q}}}$ & $\mathbb{R}^{1\times1}$ & KV-to-query head ratio
& $\mathbf{Y}_P$ & $\mathbb{R}^{\tau_{\text{in}}\times d}$ & FFN output \\
$d_{\text{kv}}=d\, R$ & $\mathbb{R}^{1\times1}$ & Key, Value dimension
& & & \\

\midrule

\multicolumn{3}{l}{\textbf{Parameters}}
&
\multicolumn{3}{@{}l}{\textbf{Decode Phase}} \\
$\mathbf{W}_{Q}$ & $\mathbb{R}^{d\times d}$ & Query weight
& $\mathbf{X}_D$ & $\mathbb{R}^{1\times d}$ & Input \\
$\mathbf{W}_K,\mathbf{W}_V$ & $\mathbb{R}^{d\times d_{\text{kv}}}$ & Key, Value weights
& $\mathbf{Q}_D,\mathbf{O}_D$ & $\mathbb{R}^{1\times d}$ & Query, Output projections \\
$\mathbf{W}_O$ & $\mathbb{R}^{d\times d}$ & Output weight
& $\mathbf{K}_D,\mathbf{V}_D$ & $\mathbb{R}^{1\times d_{\text{kv}}}$ & Key, Value projections \\
$\mathbf{W}_1$ & $\mathbb{R}^{d\times4d}$ & FFN expansion
& $\mathbf{K}_{D_{\text{cache}}},\mathbf{V}_{D_{\text{cache}}}$ & $\mathbb{R}^{(\tau_{\text{in}}+t-1)\times d_{\text{kv}}}$ & Key, Value cache \\  
$\mathbf{W}_2$ & $\mathbb{R}^{4d\times d}$ & FFN projection
& $\mathbf{S}_D$ & $\mathbb{R}^{1\times(\tau_{\text{in}}+t-1)}$ & Attention scores \\
& &
& $\mathbf{A}_D$ & $\mathbb{R}^{1\times d}$ & Attention output \\
& &
& $\mathbf{H}_D$ & $\mathbb{R}^{1\times4d}$ & Expanded activation \\
& &
& $\mathbf{Y}_D$ & $\mathbb{R}^{1\times d}$ & FFN output \\

\bottomrule
\end{tabular}
\label{tab:transformer_symbols}
\end{table*}
In this section, we present the analytical derivation of the Floating-Point Operations (\emph{FLOPs}) and Memory Operations (\emph{MemOPs}) of LLM inference based on a decoder-only Transformer, distinguishing the prefill and decode phases and assuming no quantization of weights or KV-cache. In the following derivations, we consider a dense LLM and a constant batch size, extending the derivation of  \textit{SweetSpot} \cite{sweetspot} to include the GQA mechanism. 

For a matrix multiplication between $\mathbf{A} \in \mathbb{R}^{m \times k}$ and $\mathbf{B} \in \mathbb{R}^{k \times n}$, we denote the number of FLOPs and MemOPs as $F$ and $M$ respectively, calculated as:
\begin{align*}
F &= 2mnk \\
M &= mk + kn + mn \quad \text{(read } \mathbf{A}\text{, read } \mathbf{B}\text{, write result)}
\end{align*}

\noindent
We neglect bias additions, layer normalization, softmax, and dropout, as their cost is negligible relative to projections and FFN layers. Moreover, we treat the FFN expansion factor as a fixed constant (4$\times$), consistent with the original decoder-only Transformer design and still the dominant convention across dense LLM architectures, even though some models in our evaluation set adopt slightly different ratios.

Table~\ref{tab:transformer_symbols} describes the symbols we use throughout the paper. In order to model both MHA and GQA, we define the KV-to-query head ratio $R$ as the ratio between the number of KV heads $n_{\text{kv}}$ and the number of query heads $n_{\text{q}}$. We consider the KV projections as weight matrices of shape $d \times d \cdot R = d \times d_{\text{kv}}$, reflecting that in GQA only $n_{\text{kv}} \leq n_{\text{q}}$ key and value heads are materialized. For MHA, $n_{\text{kv}} = n_{\text{q}}$ and all expressions reduce to the standard form.


\subsection{Prefill FLOPs and MemOPs}
\label{sec:prefill_flops}

During prefill, the full input sequence of length $\tau_{\text{in}}$ is processed in parallel.

\paragraph{\textbf{Attention Projections}}
$Q_P$ projection $X_PW_Q$:
\begin{flalign} \hspace*{1.5em}
\begin{aligned}
F_{Q_P} &= 2\,\tau_{\text{in}}\,d^2
\\
M_{Q_P} &= \tau_{\text{in}} d + d^2 + \tau_{\text{in}} d = 2\tau_{\text{in}} d + d^2 
\end{aligned}
\label{eq:1_prefill_proj}
&& \end{flalign}

\noindent
$K_P,V_P$ projections $X_PW_K$ and $X_PW_V$:
\begin{flalign} \hspace*{1.5em}
\begin{aligned}
F_{K_P} =  F_{V_P} &= 2\,\tau_{\text{in}}\,d^2\,R 
\\
M_{K_P} = F_{V_P} &= \tau_{\text{in}} d + d^2 R + \tau_{\text{in}} d R 
\end{aligned}
&& \end{flalign}

\noindent
Output projection $X_PW_O$:
\begin{flalign} \hspace*{1.5em}
\begin{aligned}
F_{O_P} &= 2\,\tau_{\text{in}}\,d^2 
\\
M_{O_P} &= 2\tau_{\text{in}} d + d^2 
\end{aligned}
&& \end{flalign}

\noindent
Total prefill projection complexity:
{\small
\begin{flalign} \hspace*{1.5em}
\begin{aligned}
F_{P_{\mathrm{proj}}} &= 2\tau_{\text{in}}d^2\!\left(2 + 2R\right) 
                  = 4\tau_{\text{in}}d^2\!\left(1 + R\right) 
\\
M_{P_{\mathrm{proj}}}
  &= \tau_{\text{in}} d \left(6 + 2R\right)
   + d^2 \left(2 + 2R\right) 
   \notag
\end{aligned}
&& \end{flalign}
}

\paragraph{\textbf{Attention Scores}}

\noindent
Attention scores $Q_PK_P^T$:
\begin{flalign} \hspace*{1.5em}
\begin{aligned}
F_{S_P} &= 2\,n_{\text{q}}\,\tau_{\text{in}}^2\,\frac{d}{n_{\text{q}}}
        = 2\,\tau_{\text{in}}^2\,d 
\\
M_{S_P}
  &= \tau_{\text{in}} d
  + \tau_{\text{in}} d R
  + \tau_{\text{in}}^2\, n_{\text{q}} 
\end{aligned}
&&
&& \end{flalign}

\noindent
Attention output $S_PV_P$ over $n_{\text{q}}$ heads:
\begin{flalign} \hspace*{1.5em}
\begin{aligned}
F_{A_P} &= 2\,\tau_{\text{in}}^2\,d
\\
M_{A_P}
  &= \tau_{\text{in}}^2\, n_{\text{q}}
  + \tau_{\text{in}} d R
  + \tau_{\text{in}} d 
\end{aligned}
&& \end{flalign}

\paragraph{\textbf{Feed-Forward Network}}

\noindent
FFN output $A_PW_1$ and $H_PW_2$:
\begin{flalign} \hspace*{1.5em}
\begin{aligned}  
F_{Y_P} &= 2 \times 2\,\tau_{\text{in}}\,d\cdot 4d = 16\,\tau_{\text{in}}\,d^2 
\\
M_{Y_P}
  &= \tau_{\text{in}} d + 4d^2 + 4\tau_{\text{in}} d
   + 4\tau_{\text{in}} d + 4d^2 + \tau_{\text{in}} d 
   \\
  &= 10\,\tau_{\text{in}} d + 8\,d^2
\end{aligned}
\label{eq:6_prefill_ffn}
&& \end{flalign}

\paragraph{\textbf{Total OPs}}
\label{subsec:total_prefill}
\noindent
With $L$ layers, the total prefill FLOPs is:

{\smaller
\begin{flalign} \hspace*{1.5em}
\begin{aligned}
F_{\mathrm{prefill}} 
 &= L\!\left[F_{P_{\mathrm{proj}}} + F_{S_P} + F_{A_P} + F_{Y_P}\right]  
 \\
 &= L\!\left[
  4\, \tau_{\text{in}}\,d^2\!\left(5+R\right)
  + 4\tau_{\text{in}}^2 d \right]
\end{aligned}
\label{eq:f_prefill}
&& \end{flalign}
}

\noindent
The total prefill MemOPs is:

{\smaller
\begin{flalign} \hspace*{1.5em}
\begin{aligned}
M_{\mathrm{prefill}} 
&= L\!\left[M_{P_{\mathrm{proj}}} + M_{S_P} + M_{A_P} + M_{Y_P}\right]  
 \\
&= L\left[
  \tau_{\text{in}} d \left(18 + 4R\right)
  + d^2 \left(10 + 2R\right)
  + 2\,\tau_{\text{in}}^2\, n_{\text{q}} \right]
\end{aligned}
\label{eq:m_prefill}
&& \end{flalign}
}


\subsection{Decode FLOPs and MemOPs}
\label{sec:decode_flops}
\noindent
At each decode step $t\in[1, \ldots, \tau_{\text{out}}]$, one new token is generated autoregressively. The KV-cache at step $t$ holds ($\tau_{\text{in}} + t - 1$)
tokens: the $\tau_{\text{in}}$ input tokens and $t-1$ previously generated tokens. 

\paragraph{\textbf{Attention Projections}}
\noindent
$Q_D$ projection $X_DW_Q$: 
\begin{flalign} \hspace*{1.5em}
\begin{aligned}
F_{Q_D} &= 2\,d^2 
\\
M_{Q_D} &= d + d^2 + d = 2d + d^2 
\end{aligned}
\label{eq:9_decode_proj}
&& \end{flalign}

\noindent
$K_D,V_D$ projection $X_DW_K$ and $X_DW_V$:
\begin{flalign} \hspace*{1.5em}
\begin{aligned}
F_{K_D} = F_{V_D} &= 2\,d^2\,R
\\
M_{K_D} = F_{V_D} &= d + d^2 R + d R 
\end{aligned}
&& \end{flalign}

\noindent
Output projection $X_DW_O$:
\begin{flalign} \hspace*{1.5em}
\begin{aligned}
F_{O_D} &= 2\,d^2 
\\
M_{O_D} &= 2d + d^2 
\end{aligned}
&& \end{flalign}

\noindent
Total decode projection complexity:
{\small
\begin{flalign} \hspace*{1.5em}
\begin{aligned}
F_{D_{\mathrm{proj}}} &= 2d^2\left(2 + 2 R\right)
                  = 4d^2\!\left(1 + R\right)
\\
M_{D_{\mathrm{proj}}}
  &= d \left(6 + 2 R\right)
   + d^2 \left(2 + 2 R\right) 
   \notag 
\end{aligned}
&& \end{flalign}
}

\paragraph{\textbf{Attention Scores}}

\noindent
Attention scores $Q_DK_{D_{\mathrm{cache}}}^T$:
\begin{flalign} \hspace*{1.2em}
\begin{aligned}
F_{S_D}(t) &= 2(\tau_{\text{in}} + t - 1)d 
\\
M_{S_D} (t)
  &= d
  + (\tau_{\text{in}}+t-1)\, d R
  + (\tau_{\text{in}}+t-1)\, n_{\text{q}} 
\end{aligned}
&& \end{flalign}

\noindent
Attention output $S_DV_{D_{\mathrm{cache}}}$ over $n_{\text{q}}$ heads:
\begin{flalign} \hspace*{1.2em}
\begin{aligned}
    F_{A_D}(t) &= 2(\tau_{\text{in}} + t - 1)d 
\\
M_{A_D}(t)
  &= (\tau_{\text{in}}+t-1)\, n_{\text{q}}
  + (\tau_{\text{in}}+t-1)\, d R
  + d 
\end{aligned}
&& \end{flalign}

\paragraph{\textbf{Feed-Forward Network}}

\noindent
FFN output $A_DW_1$ and $H_DW_2$:
\begin{flalign} \hspace*{1.2em}
\begin{aligned}
F_{Y_D}(t) &= 16d^2 
\\
M_{Y_D}(t)
  &= d + 4d^2 + 4d + 4d + 4d^2 + d \\
  &= 10\,d + 8\,d^2 
\end{aligned}
\label{eq:14_decode_ffn}
&& \end{flalign}

\paragraph{\textbf{Total OPs}}
\label{subsec:total_decode}
\noindent
Using the closed-form sum: 
\begin{flalign} \hspace*{1.5em}
\begin{aligned}
\sum^{\tau_{\text{out}}}_{t=1} t = \frac{\tau_{\text{out}}(\tau_{\text{out}} + 1)}{2}
\notag
\end{aligned}
\end{flalign}
\noindent
and summing over $\tau_{\text{out}}$ generated tokens and $L$ layers, we obtain the total decode FLOP count:

{
\smaller
\begin{flalign} \hspace*{1em}
\begin{aligned}
F_{\mathrm{decode}} 
&= L\sum^{\tau_{\text{out}}}_{t=1}\!\left[F_{D_{\mathrm{proj}}} + F_{S_D}(t) + F_{A_D}(t) + F_{Y_D}(t) \right] \\
&= L\sum^{\tau_{\text{out}}}_{t=1}\!\left[
  4d^2\!\left(5 + R\right) + 4d(\tau_{\text{in}} + t - 1)
\right] \\
&=L\left[4d^2\tau_{\text{out}}\left(5 + R\right) + 4d\left(\tau_{\text{out}} \tau_{\text{in}} + \frac{\tau_{\text{out}}(\tau_{\text{out}}-1)}{2}\right)\right] 
\end{aligned}
\label{eq:f_decode} 
&& \end{flalign}

}

\noindent
Similarly, we obtain the total decode MemOPs:

{
\smaller
\begin{flalign} \hspace*{1em}
\begin{aligned}
M_{\text{decode}}
&= L\sum^{\tau_{\text{out}}}_{t=1}\!\left[M_{D_{\mathrm{proj}}} + M_{S_D} + M_{A_D} + M_{Y_D} \right] \\
  &= L\Bigg[
  \tau_{\text{out}}\, d\left(18 + 2R\right)
  + \tau_{\text{out}}\, d^2\left(10 + 2 R\right) \\
  &\quad + 2\left(n_{\text{q}} + dR\right)
  \left(\tau_{\text{out}}\,\tau_{\text{in}} + \frac{\tau_{\text{out}}(\tau_{\text{out}} - 1)}{2}\right)
  \Bigg]
\end{aligned}
\label{eq:m_decode}
&& \end{flalign}
}

\section{SIMPLE Derivation}
\label{sec:model_derivation}
In this section, we present the formulation of \textit{SIMPLE}, the proposed analytical model for predicting the energy efficiency of LLM inference, for a generic LLM architecture $f(d, L, n_{\text{q}}, n_{\text{kv}})$ and the input/output sequence $f(\tau_{\text{in}}, \tau_{\text{out}})$. 
The model is built on the complexity analysis introduced in the previous section, more precisely on Equations~(\ref{eq:1_prefill_proj}--\ref{eq:14_decode_ffn}). 
These equations offer different valid groupings of their inner terms according to the number of fitting coefficients, each implying a different interpretation for each coefficient.

A first approach we propose is to assign a meaning to each coefficient in terms of Joule/FLOPs or Joule/MemOPs.  We refer to these models as \emph{OPs-centric}. 
A second orthogonal approach is to drop the interpretation but preserve only the relational one with respect to $f(\tau_{\text{in}}, \tau_{\text{out}})$. We refer to these models as \emph{IO-centric}. In these formulations, we include the LLM architectural hyperparameters ($d, L, n_{\text{q}},n_{\text{kv}}$) in order to obtain LLM-agnostic models. While the OPs-centric models serve as an exploratory baseline, \textit{SIMPLE} is ultimately derived within the IO-centric family.

The proposed models are summarized in Table~\ref{tab:models}. As direct comparisons, we include the $E_{\text{tok}}$ formulations of \textit{SweetSpot}~\cite{sweetspot} and the two LLM-agnostic models from related work: Vartziotis et al.~\cite{vartziotis}, which can be classified as OPs-centric as it explicitly assigns meaning to its terms, and Zhan et al.~\cite{zhan}, which falls among the IO-centric models.
\begin{table*}
\centering
\caption{Summary of the proposed energy models and selected baselines for predicting the energy per generated token, $E_{\text{tok}}$. \textit{SweetSpot} is included for direct comparison. For the LLM-agnostic \textit{SweetSpot} model, $\phi_0,\ldots,\phi_5$ denote architecture-dependent analytical coefficients, whose explicit expressions are given in Equation~\ref{eq:sweetspot_extended}.}
\label{tab:models}
\footnotesize

\renewcommand{\arraystretch}{1.6}
\setlength{\tabcolsep}{0pt}
\setlength{\aboverulesep}{2mm}
\setlength{\belowrulesep}{2mm}

\begin{tabular}{@{}m{0.15\linewidth}@{\hspace{0.04\linewidth}}m{0.19\linewidth}@{\hspace{0.01\linewidth}}>{\centering\arraybackslash}m{0.07\linewidth}@{\hspace{0.04\linewidth}}m{0.40\linewidth}@{}}
\toprule
\textbf{Approach} & \textbf{Model} & \textbf{\# coeff.} & \textbf{Energy per token, $E_{\text{tok}}$} \\
\midrule

\multirow{4}{*}[-3em]{OPs-centric}
& \makecell[l]{Vartziotis et al. \cite{vartziotis}}
& 6
& $\dfrac{\alpha_{\text{TC}}C_{\text{TC}}}{\tau_{\text{out}}}
+\dfrac{e_{\text{HBM}}\,B\,\eta(N)}{\tau_{\text{out}}}$ \\
\addlinespace[4mm]

& \makecell[l]{FLOPs-MemOPs}
& 3
& $\theta_0
+\theta_1\dfrac{F_{\text{prefill}}+F_{\text{decode}}}{\tau_{\text{out}}}
+\theta_2\dfrac{M_{\text{prefill}}+M_{\text{decode}}}{\tau_{\text{out}}}$ \\
\addlinespace[4mm]

& \makecell[l]{Prefill-Decode}
& 5
& $\theta_{0}
+\theta_{1}\dfrac{F_{\text{prefill}}}{\tau_{\text{out}}}
+\theta_{2}\dfrac{F_{\text{decode}}}{\tau_{\text{out}}}
+\theta_{3}\dfrac{M_{\text{prefill}}}{\tau_{\text{out}}}
+\theta_{4}\dfrac{M_{\text{decode}}}{\tau_{\text{out}}}$ \\
\addlinespace[4mm]

& \makecell[l]{Fine-Grained}
& 25
& $\theta_0
+\sum_{i=1}^{24}\theta_i\dfrac{L\,C_i}{\tau_{\text{out}}} \ \qquad
C_i \in \left\{ F_{Q_P}, \ldots, \sum_{t=1}^{\tau_{\text{out}}} M_{A_D}(t) \right\}$ \\
\midrule

\multirow{4}{*}[-3em]{IO-centric}
& \makecell[l]{Zhan et al. \cite{zhan}}
& 5
& $\theta_{0}
+\theta_{1}S\dfrac{1}{\tau_{\text{out}}}
+\theta_{2}\dfrac{\tau_{\text{in}}}{\tau_{\text{out}}}
+\theta_{3}S\dfrac{\tau_{\text{in}}}{\tau_{\text{out}}}
+\theta_{4}S$ \\
\addlinespace[4mm]

& \makecell[l]{SweetSpot \cite{sweetspot}}
& 6
& $\theta_{0}
+\theta_{1}\dfrac{\tau_{\text{in}}^{2}}{\tau_{\text{out}}}
+\theta_{2}\tau_{\text{in}}
+\theta_{3}\dfrac{\tau_{\text{in}}}{\tau_{\text{out}}}
+\theta_{4}\tau_{\text{out}}
+\dfrac{\theta_{5}}{\tau_{\text{out}}}$ \\
\addlinespace[4mm]

& \makecell[l]{LLM-agnostic SweetSpot}
& 7
& $\theta_{0}
+\theta_{1}\dfrac{\tau_{\text{in}}^{2}}{\tau_{\text{out}}}\phi_0
+\theta_{2}\tau_{\text{in}}\phi_1
+\theta_{3}\dfrac{\tau_{\text{in}}}{\tau_{\text{out}}}\phi_2
+\theta_{4}\tau_{\text{out}}\phi_3
+\dfrac{\theta_{5}}{\tau_{\text{out}}}\phi_4
+\theta_{6}\phi_5$ \\
\addlinespace[4mm]

& \textbf{SIMPLE}
& \textbf{4}
& $\theta_{0} +  
\theta_{1} \, d^2 L \, 
\dfrac{\tau_{\text{in}}}{\tau_{\text{out}}}
+ \theta_{2} \, d \, L \, R
\Big(
    \tau_{\text{in}} + \dfrac{\tau_{\text{out}}}{2}
\Big)
+ \theta_3 d \, L \
$\\

\bottomrule
\end{tabular}
\end{table*}

\subsection{OPs-centric Approach}
We formulate three models that assign a meaning to each coefficient in terms of Joule/FLOPs and Joule/MemOPs at different levels of granularity.

\paragraph{\textbf{FLOPs-MemOPs}} The first formulation assigns one coefficient to Equations~(\ref{eq:f_prefill}, \ref{eq:f_decode}) and another to Equations~(\ref{eq:m_prefill}, \ref{eq:m_decode}):

{
\small
\begin{equation}
E_{\text{tok}}
=
\theta_0 + 
\theta_1 \frac{F_{\text{prefill}} + F_{\text{decode}}}{\tau_{\text{out}}} + \theta_2 \frac{ M_{\text{prefill}} + M_{\text{decode}}}{\tau_{\text{out}}}
\label{eq:hc_two_terms}
\end{equation}
}
\noindent
The model assumes that these two components contribute independently to the total inference energy, assigning a separate cost to each, regardless of whether the operations are performed in the prefill or the decode phase.

\paragraph{\textbf{Prefill-Decode}}
The second formulation differentiate these energy costs by prefill and decode phases, and assigns individual coefficients to Equations~(\ref{eq:f_prefill}, \ref{eq:f_decode}, \ref{eq:m_prefill}, \ref{eq:m_decode}), allowing the model to capture potentially different operating conditions and energy costs of the two phases:

{
\small
\begin{equation}
E_{\text{tok}}
=
\theta_0 + 
\theta_1 \frac{F_{\text{prefill}}}{\tau_{\text{out}}} + \theta_2 \frac{F_{\text{decode}}}{\tau_{\text{out}}} + \theta_3 \frac{M_{\text{prefill}}}{\tau_{\text{out}}} + \theta_4 \frac{M_{\text{decode}}}{\tau_{\text{out}}}
\label{eq:hc_split}
\end{equation}
}

\paragraph{\textbf{Fine-Grained}}
The third formulation maximizes the fitting granularity assigning a coefficient to each intermediate term, but perishing the physical meaning of Joule/FLOPs and Joule/MemOPs:

{
\small
\begin{equation}
E_{\text{tok}}
=
\theta_0
+
\sum_{i=1}^{24}
\theta_i\, \frac{L \, C_i}{\tau_{\text{out}}},
\label{eq:hc_all}
\end{equation}
}
\noindent
where:
{
\footnotesize
\begin{equation}
C_i \in \left\{ F_{Q_P}, M_{Q_P}, \ldots \sum_{t=1}^{\tau_{\text{out}}}{F_{A_D}(t)},\sum_{t=1}^{\tau_{\text{out}}}{M_{A_D}(t)} \right\}
\notag
\end{equation}
}
\noindent
denotes the complexity of each intermediate term across prefill and decode from Equations~(\ref{eq:1_prefill_proj}--\ref{eq:6_prefill_ffn}, \ref{eq:9_decode_proj}--\ref{eq:14_decode_ffn}). 

\subsection{IO-centric Approach}
The previous models explore different ways of incorporating the analytical FLOPs and MemOPs into the energy prediction model. While these formulations explicitly account for the underlying computational cost, they also introduce a growing number of terms and fitting coefficients. The IO-centric approach instead groups the complexity terms according to their dependency on $f(\tau_{\text{in}}, \tau_{\text{out}})$, simplifying or discarding those of negligible contribution.  

In the following paragraphs, we first introduce \emph{LLM-agnostic SweetSpot}, a natural extension of \emph{SweetSpot}, \cite{sweetspot} used as a comparison baseline, then present \textit{SIMPLE}. 

\paragraph{\textbf{LLM-agnostic SweetSpot}}
The \textit{SweetSpot} model \cite{sweetspot} captures quadratic dependencies on $f(\tau_{\text{in}}, \tau_{\text{out}})$, but is fitted for individual LLMs.
All architectural information is implicitly absorbed into the learned coefficients, making it impossible to generalize across different LLM architectures.
A natural extension is therefore to preserve \textit{SweetSpot}'s
functional dependencies on $f(\tau_{\text{in}}, \tau_{\text{out}})$ while making the coefficients explicitly dependent on the LLM architecture $f(d, L, n_{\text{q}}, n_{\text{kv}})$. To this end, we group the complexity terms from Equations~(\ref{eq:f_prefill}, \ref{eq:m_prefill},  \ref{eq:f_decode}, \ref{eq:m_decode})  to match the corresponding terms of the \textit{SweetSpot} formulation reported in Table~\ref{tab:models}. 

{
\smaller
\begin{equation}
\label{eq:sweetspot_extended}
\begin{aligned}
E_{\text{tok}}=\
& \theta_{0} + \theta_{1}\, \frac{\tau_{\text{in}}^2}{\tau_{\text{out}}} \, L \left(4d + 2n_{\text{q}}\right) + \theta_{2}\, \tau_{\text{in}} \, L \left(4d + 2dR + 2n_{\text{q}}\right) \\
+\ & \theta_{3}\, \frac{\tau_{\text{in}}}{\tau_{\text{out}}} \, L \left[ d^2 \left(4R + 20\right) + d \left(4R + 18\right) \right]  \\
+\ & \theta_{4}\, \tau_{\text{out}} \, L \left(2d + dR + n_{\text{q}}\right)
+ \frac{\theta_{5}}{\tau_{\text{out}}} \, L \, d^2 \left(2R + 10\right) \\
+\ & \theta_6 L \left[ d^2 \left(6R + 30\right) + d \left(2R + 18\right) - 2d - \left(dR + n_{\text{q}}\right) \right] \\
\end{aligned}
\end{equation}
}
\vspace{0.2em}
\noindent
The resulting formulation preserves the \textit{SweetSpot} dependencies while explicitly incorporating the architectural complexity of the target LLM, thereby enabling generalization across multiple LLM architectures. 
However, despite sharing a similar number of fitting coefficients and functional dependencies with \textit{SweetSpot}, this formulation involves a considerable number of terms, similarly to the previously derived \textit{Fine-Grained} model, since it groups complexity terms without assessing their individual significance or exploring optimal grouping strategies. As a consequence, it provides limited insights into how each architectural parameter and inference phase contributes to energy consumption.

\paragraph{\textbf{SIMPLE}}
To obtain a compact, interpretable, and generalizable model, we systematically evaluated different simplification and term-aggregation strategies, starting from Equations~({\ref{eq:f_prefill}, \ref{eq:m_prefill}, \ref{eq:f_decode}, \ref{eq:m_decode}}). In particular, we adopted three complementary methods to refine the model. First, we performed term ablations, removing inner terms with low statistical significance. Second, we simplified the resulting expressions by discarding lower-order terms. Third, we explored different grouping strategies for the remaining terms. Based on these refinement methods, we derived the following $E_{\text{tot}}$ formulation of \textit{SIMPLE}:

\newcommand{\bracestrut}{\vphantom{\Big(\frac{\tau_{\text{out}}^2}{2}\Big)}}

{
\small
\begin{equation}
\label{eq:sweetspotv2_etot}
E_{\text{tot}}
=
\underbrace{\theta_{1} \, d^2 L\, \tau_{\text{in}} \bracestrut}_{\text{\small Prefill}}
+ \underbrace{\overbrace{\theta_{2} d \, L \, R
\Big(
    \tau_{\text{in}} \tau_{\text{out}} + \frac{\tau_{\text{out}}^2}{2}
        \bracestrut}^{\text{\small KV-cache}}
\Big)
+ \overbrace{(\theta_0 + \theta_3 d \, L \,) \tau_{\text{out}} \bracestrut}^{\text{\small Next-token}}
\bracestrut}_{\text{\small Decode}}
\notag
\end{equation}
}

\noindent
where the total energy consumption combines the cost of the \emph{Prefill} and \emph{Decode} phases. 

The \emph{Prefill} term
accounts for processing the input prompt and is obtained by refining Equations~(\ref{eq:f_prefill}, ~\ref{eq:m_prefill}). It scales linearly with the input length $\tau_{\text{in}}$ and the number of layers $L$, and quadratically with the hidden dimension $d$.
The \emph{Decode} component covers the generation of the output tokens and is obtained by refining Equations~(\ref{eq:f_decode}, \ref{eq:m_decode}). It comprises two terms. The \emph{KV-cache} term 
captures the cost of reading the KV-cache, which grows as the context lengthens with each generated token. 
The \emph{Next-token} term
represents the cost that accumulates at a constant rate per generated token. Both terms scale linearly with $d$ and $L$, while the \emph{KV-cache} term additionally scales with the KV-to-query head ratio $R$. 

The fitting coefficients $\theta_i$ weigh each term independently, assigning a distinct cost to each phase and separating the \emph{KV-cache} cost from the \emph{Next-token} cost.
Dividing
by the number of generated tokens $\tau_{\text{out}}$ yields the
per-token formulation $E_{\text{tok}}$ of \textit{SIMPLE}:

{
\small
\begin{equation}
\label{eq:sweetspotv2}
E_{\text{tok}}
=
\underbrace{%
  \termbox{prefillcol}{\theta_{1}\, d^2 L\, \frac{\tau_{\text{in}}}{\tau_{\text{out}}}}%
}_{\mathclap{\text{\small Prefill Amortization}}}
+
\underbrace{%
  \overbrace{%
    \termbox{kvcol}{\theta_{2}\, d\, L\, R \left( \tau_{\text{in}} + \frac{\tau_{\text{out}}}{2} \right)}%
  }^{\mathclap{\text{\small KV-cache}}}
  +
  \overbrace{%
    \termbox{nexttokcol}{\theta_{0} + \theta_{3}\, d\, L}%
  }^{\mathclap{\text{\small Next-token}}}
}_{\mathclap{\text{\small Decode Accumulation}}}
\end{equation}
}

\noindent
The Prefill term becomes a \emph{Prefill Amortization} term, reflecting 
how the fixed cost of processing the prompt is spread over the generated tokens.  The Decode term becomes a \emph{Decode Accumulation} term, in which the KV-cache cost grows linearly with the context length, and the Next-token cost reduces to a constant per generated token. An interactive demo of \textit{SIMPLE} is publicly available.~\footnote{\href{https://huggingface.co/spaces/ecs-lab/SIMPLE}{ecs-lab/SIMPLE}}

\section{Experimental Validation}
\label{sec:experimental_study}
This section details the experimental study conducted to validate \textit{SIMPLE}.
Section \ref{sec:dataset_setup} describes the dataset, available on Zenodo~\footnote{\href{https://doi.org/10.5281/zenodo.18714476}{doi.org/10.5281/zenodo.18714476}}, introduced in~\cite{sweetspot} and reused in our experiments, summarizing its key observations  on energy consumption, namely the relevance of the attention mechanism and the non-linear relationship with input and output lengths. 
Section \ref{subsubsec:analysis} presents our results. The code of the experiments is available at a public repository.~\footnote{\href{https://gitlab.com/ecs-lab/simple-law}{gitlab.com/ecs-lab/simple-law}}

\subsection{Dataset Description}
\label{sec:dataset_setup}
The dataset contains energy measurements of TensorRT-LLM inference traces collected from a diverse set of widely used dense LLMs running on NVIDIA H100 accelerators. 
The included models span a range of LLM sizes and architectures, including MHA and on GQA with different KV grouping ratios, using a standard bf16 format for weights and KV-cache.
The full list of models is given in Table~\ref{tab:llm_set}.

The dataset comprises synthetic requests processed using \texttt{trtllm-bench} configured in \textit{throughput} mode, with request concurrency set to 1000 and a maximum batch size of 1024. The maximum sequence length and the maximum number of batched tokens are both set to 8192.
The experiments were conducted using one GPU per run on a server equipped with an Intel Xeon Platinum 8480+ CPU and four NVIDIA H100 Tensor Core GPUs connected via NVLink. 
Input and output lengths independently span 64--4096 tokens in powers of two, yielding 49 combinations with 1000 requests each. 
GPU power was sampled every 500~ms using \texttt{pynvml}~\cite{pynvml}.
The per-query energy was obtained by multiplying the average GPU power over 1000-request by the total measurement time, and dividing by the number of requests.
Since neither the baseline models nor the SIMPLE model explicitly accounts for batching, we use these fixed settings, 
and fit energy per individual request.

\begin{table}
\footnotesize
\centering
\caption{LLM test set grouped into size-based families (1B–9B parameters), with architectural hyperparameters for each model.}
\begin{tabularx}{\columnwidth}{lllllll}
\toprule
\textbf{Name} & \textbf{Size (B)} & \textbf{Attention} & $\mathbf{n_{\text{q}}}$ & $\mathbf{R}$ & $\mathbf{d}$ & $\mathbf{L}$ \\
\midrule
Llama 3.2   & 1   & GQA & 32 & 1/4 & 2048 & 16 \\
OPT         & 1.3 & MHA & 32 & 1   & 2048 & 24 \\
Qwen 2      & 1.5 & GQA & 12 & 1/6 & 1536 & 28 \\

Gemma 2     & 2   & GQA & 8  & 1/2 & 2304 & 26 \\
OPT         & 2.7 & MHA & 32 & 1   & 2560 & 32 \\
Llama 3.2   & 3   & GQA & 24 & 1/3 & 3072 & 28 \\
Granite     & 3   & MHA & 32 & 1   & 2560 & 32 \\

OPT         & 6.7 & MHA & 32 & 1   & 4096 & 32 \\
Qwen 2      & 7   & GQA & 28 & 1/7 & 3584 & 28 \\
Falcon-RW   & 7   & MHA & 64 & 1   & 4096 & 36 \\
Granite     & 8   & GQA & 32 & 1/4 & 4096 & 36 \\
Llama 3.1   & 8   & GQA & 32 & 1/4 & 4096 & 32 \\
Gemma 2     & 9   & GQA & 16 & 1/2 & 3584 & 42 \\
\bottomrule
\label{tab:llm_set}
\end{tabularx}
\end{table}
 
\begin{figure}
\centering
\includegraphics[width=\columnwidth]{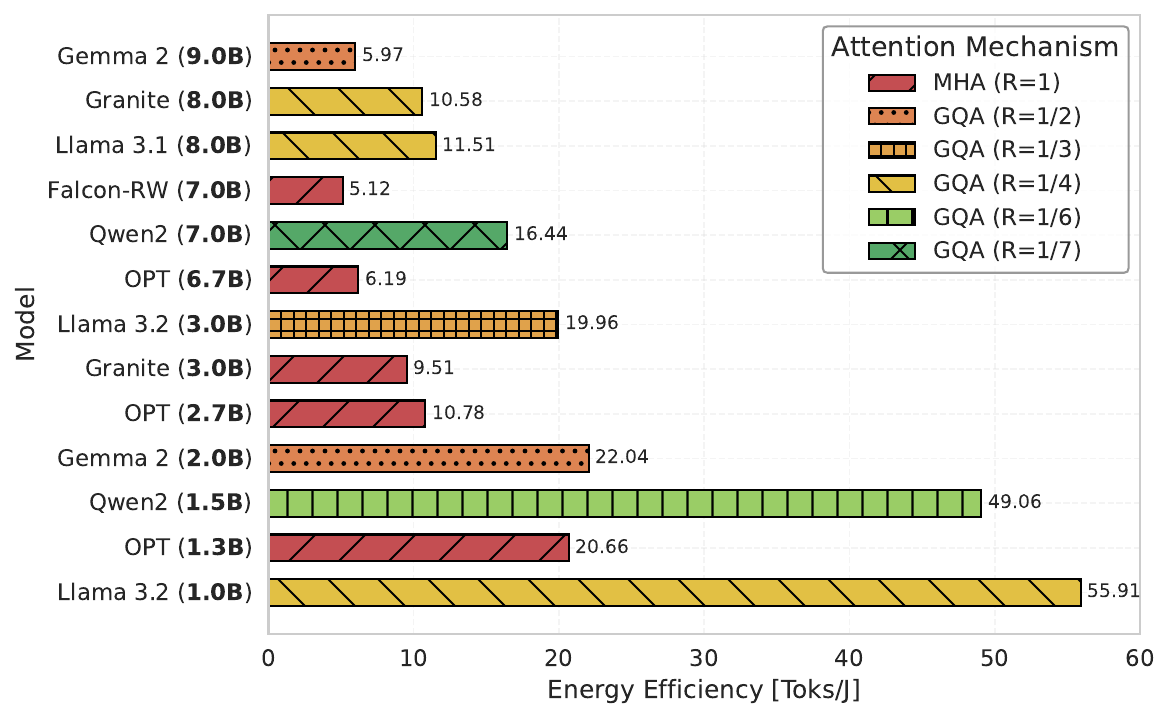}
\caption{Average energy efficiency across the evaluated LLMs, highlighting the adopted attention architecture.}
\label{fig:average_efficiency}  
\end{figure}

Figure~\ref{fig:average_efficiency} reports the average tokens per joule computed on the 49 different workload configurations across the different LLMs in the dataset. 
Smaller models and GQA-based architectures generally achieved higher efficiency (Figure~\ref{fig:average_efficiency}): Llama 3.2 1B reached 55.91~tok/J, compared with 5.12~tok/J for Falcon-RW 7B. 
Among similarly sized models, Qwen2 1.5B (GQA R=1/7) achieved 49.06~tok/J versus 20.66~tok/J for OPT 1.3B (MHA R=1). 
This underlines the importance of modeling the attention architecture alongside other LLM architectural hyperparameters. 
Model size alone is not enough to characterize inference energy consumption. 

\begin{figure}
\centering
\includegraphics[width=0.7\columnwidth]{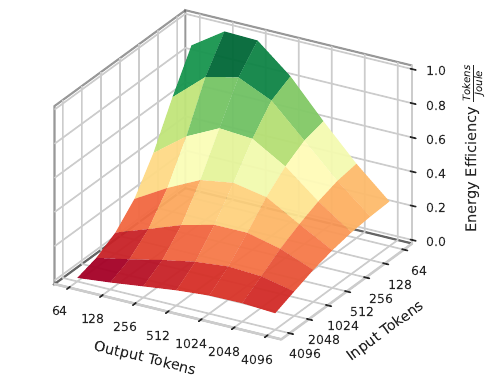}
\caption{Aggregated energy efficiency across the dataset as a function of input and output sequence length.}
\label{fig:heatmap_average}  
\end{figure}

Figure~\ref{fig:heatmap_average} reports an aggregated average of the tokens per joule computed on the different models across the 49 different workload configurations in the dataset. 
Efficiency generally peaked at short inputs ($\tau_{\text{in}}=64$) and intermediate outputs ($\tau_{\text{out}}\in[128,256]$), and was lowest for long inputs and short outputs. 
Across models, the best configuration was $(33.41\pm4.90)\times$ more efficient than the worst. 
These results showed a model-dependent energy-efficiency ``sweet spot" associated with model size, attention architecture, and sequence lengths.
These findings provide the basis for the extended analysis presented here.


\subsection{Experimental Results}
\label{subsubsec:analysis}
In this section, we evaluate the accuracy of the proposed \textit{SIMPLE} model in predicting the energy per token against that of the SOTA models and the other OPs-centric and IO-centric models reported in Table~\ref{tab:models}. All models have been fitted using Weighted Least Squares (WLS), since the variance of the measured token energy increases with the number of processed tokens, indicating heteroscedasticity.
\begin{table*}
\centering
\caption{
Comparison of energy models across three evaluation
protocols: LOOCV, $k$-fold Cross Validation (CV), and the full-dataset fit, using WLS ($w=1/y^2$). For
cross-validation protocols, we report mean $R^2$ and MAPE, $\pm$ standard deviation across folds.
}
\label{tab:cv_loocv_kfold}
\smaller

\makebox[\textwidth][c]{%
\begin{tabular}{lcccccccc}
\toprule
 & \multicolumn{4}{c}{\textbf{OPs-centric}}
 & \multicolumn{4}{c}{\textbf{IO-centric}} \\
\cmidrule(lr){2-5} \cmidrule(lr){6-9}
 & \makecell[c]{Vartziotis~et~al. \\ \smaller \cite{vartziotis}}
 & \makecell[c]{FLOPs\\MemOPs}
 & \makecell[c]{Prefill\\Decode}
 & \textbf{\makecell[c]{Fine\\Grained}}
 & \makecell[c]{Zhan~et~al.\\ \smaller \cite{zhan}}
 & \makecell[c]{SweetSpot\\ \smaller \cite{sweetspot}}
 & \makecell[c]{LLM-agnostic\\ SweetSpot}
 & \textbf{\makecell[c]{SIMPLE}}
 \\
\midrule
\textbf{\# coefficients} & 6 & 3 & 5 & \textbf{25} & 5 & 6 & 7 & \textbf{4} \\

\midrule
\multicolumn{9}{l}{\textbf{LOOCV}} \\
$R^2$
& \makecell{$0.645$\\{\scriptsize $(\pm 0.153)$}}
& \makecell{$0.551$\\{\scriptsize $(\pm 0.348)$}}
& \makecell{$0.522$\\{\scriptsize $(\pm 0.385)$}}
& \makecell{$\mathbf{0.945}$\\{\scriptsize ($\pm 0.046$)}}
& \makecell{$0.456$\\{\scriptsize $(\pm 0.387)$}}
& \makecell{$-0.119$\\{\scriptsize $(\pm 0.858)$}}
& \makecell{$0.780$\\{\scriptsize $(\pm 0.175)$}}
& \makecell{$\mathbf{0.944}$\\{\scriptsize ($\pm 0.057$)}}
\\

MAPE (\%)
& \makecell{$37.743$\\{\scriptsize $(\pm 14.357)$}}
& \makecell{$40.392$\\{\scriptsize $(\pm 10.793)$}}
& \makecell{$37.525$\\{\scriptsize $(\pm 14.091)$}}
& \makecell{$\mathbf{14.208}$\\{\scriptsize ($\pm 6.826$)}}
& \makecell{$44.962$\\{\scriptsize $(\pm 13.935)$}}
& \makecell{$69.444$\\{\scriptsize $(\pm 23.455)$}}
& \makecell{$30.390$\\{\scriptsize $(\pm 9.393)$}}
& \makecell{$\mathbf{13.252}$\\{\scriptsize ($\pm 5.298$)}}
\\

\midrule
\multicolumn{9}{l}{\textbf{$k$-fold CV}} \\
$R^2$
& \makecell{$0.679$\\{\scriptsize $(\pm 0.131)$}}
& \makecell{$0.505$\\{\scriptsize $(\pm 0.227)$}}
& \makecell{$0.513$\\{\scriptsize $(\pm 0.215)$}}
& \makecell{$\mathbf{0.967}$\\{\scriptsize ($\pm 0.013$)}}
& \makecell{$0.444$\\{\scriptsize $(\pm 0.245)$}}
& \makecell{$-0.024$\\{\scriptsize $(\pm 0.151)$}}
& \makecell{$0.762$\\{\scriptsize $(\pm 0.108)$}}
& \makecell{$\mathbf{0.965}$\\{\scriptsize ($\pm 0.015$)}}
\\

MAPE (\%)
& \makecell{$37.734$\\{\scriptsize $(\pm 2.703)$}}
& \makecell{$39.510$\\{\scriptsize $(\pm 3.245)$}}
& \makecell{$34.929$\\{\scriptsize $(\pm 3.690)$}}
& \makecell{$\mathbf{10.895}$\\{\scriptsize ($\pm 1.051$)}}
& \makecell{$41.707$\\{\scriptsize $(\pm 3.539)$}}
& \makecell{$63.269$\\{\scriptsize $(\pm 2.607)$}}
& \makecell{$27.694$\\{\scriptsize $(\pm 2.590)$}}
& \makecell{$\mathbf{12.038}$\\{\scriptsize ($\pm 0.896$)}}
\\

\midrule

\multicolumn{9}{l}{\textbf{Full-dataset Fit}} \\
$R^2$
& $0.725$
& $0.581$
& $0.585$
& $\mathbf{0.970}$
& $0.530$
& $0.033$
& $0.799$
& $\mathbf{0.969}$
\\

MAPE (\%)
& $37.740$
& $39.384$
& $34.748$
& $\mathbf{10.658}$
& $41.438$
& $62.665$
& $27.443$
& $\mathbf{11.974}$
\\

\bottomrule
\end{tabular}%
}

\end{table*}
\begin{figure*}[t]
  \centering
  \includegraphics[width=\textwidth]{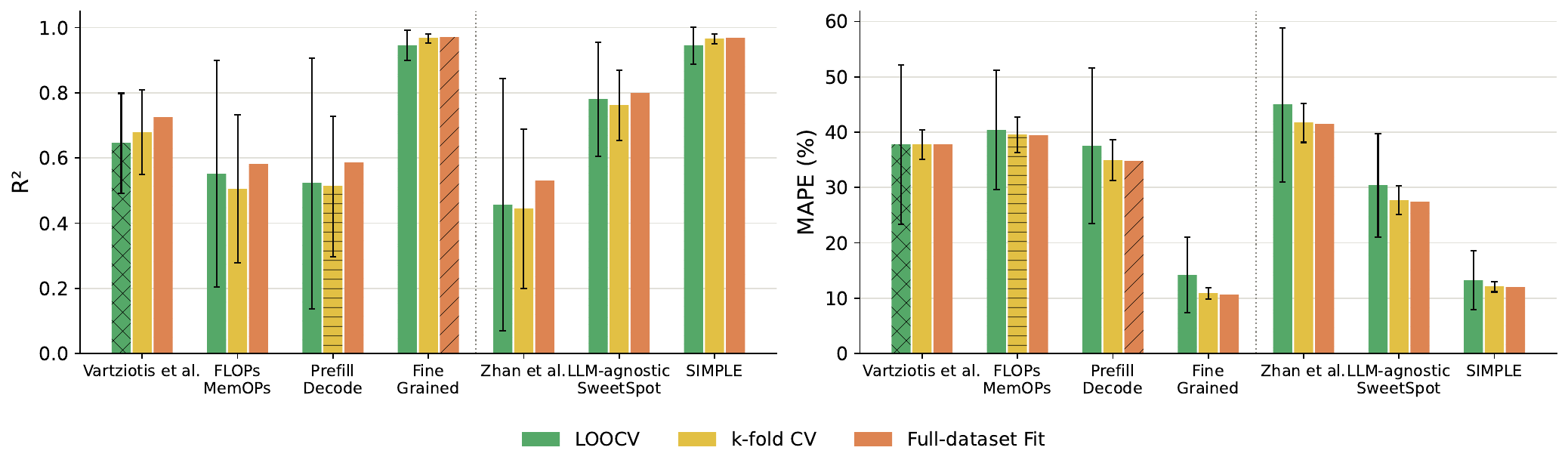}
  \caption{
  Graphical comparison of the results in Table~\ref{tab:cv_loocv_kfold}. Error bars denote standard deviation across folds. }
  \label{fig:cv_loocv_kfold}
\end{figure*}

Table~\ref{tab:cv_loocv_kfold} reports the results of this comparison across three evaluation protocols: (i) Leave-One-Out Cross Validation (LOOCV), which evaluates generalization to previously unseen LLM architectures by withholding the corresponding architectural hyperparameters ($d$, $L$, $n_{\text{q}}$, $n_{\text{kv}}$) during fitting; (ii) $k$-fold cross validation, which evaluates model stability and robustness using 10-fold cross validation with 5 repetitions over individual $(t_{\mathrm{in}}, t_{\mathrm{out}})$ observations randomly shuffled across the complete dataset, ensuring each fold contains samples from all 13 architectures, in both the training and test sets,
and (iii) full-dataset fit, used to generate the final fitting coefficients for energy estimation. In the table, columns correspond to the fitted models and rows report the accuracy results under each protocol. Figure~\ref{fig:cv_loocv_kfold} shows the values reported in Table~\ref{tab:cv_loocv_kfold}, except for \textit{SweetSpot}, whose scale would otherwise compress the remaining variants, at the expense of readability.
 We then use the full-dataset fit coefficients to compare \textit{SIMPLE} against LLM-specific models on the same dataset. 
For each model and validation protocol, we report the coefficient of determination ($R^2$) and the Mean Absolute Percentage Error (MAPE).
All three protocols show similar comparative results.

\paragraph{\textbf{Cross Validation}}
When comparing LOOCV results, SOTA approaches of Zhan et al.~\cite{zhan} and Vartziotis et al.~\cite{vartziotis} achieve $R^2$ values of $0.456$ and $0.645$, respectively. The proposed \textit{LLM-agnostic SweetSpot} outperforms SOTA models, reaching an $R^2 $ of $0.780$, showing that extending the previously proposed \textit{SweetSpot} formulation with LLM architectural hyperparameters makes it LLM-agnostic, able to predict across LLMs without separate per-model fits, improving MAPE from 69.444 to 30.390.
Despite being more accurate than SOTA models, its accuracy is far from optimal.
Among OPs-centric models, \textit{Fine-Grained} is the most accurate, reaching $R^2 = 0.945$, while \textit{FLOPs-MemOPs} and \textit{Prefill-Decode} obtain substantially lower $R^2$ of $0.551$ and $0.522$, respectively. 
This shows that considering only one coefficient per operation (FLOPs and MemOPs) or
per inference phase lowers the attainable accuracy, which drops by $2.7\times$ in MAPE compared to \textit{Fine-Grained}.

The proposed \textit{SIMPLE} model, which groups fine-grained terms into an interpretable formulation consisting of the terms corresponding to Prefill, KV-cache and Next-token costs, reduces the fitting coefficients by $6.25\times$ and
achieves the highest accuracy, with an $R^2$ of $0.944$ and a MAPE of $13.252\%$, compared to $0.945$ and $14.208\%$ for the \textit{Fine-Grained} model, while compared to the SOTA models, \textit{SIMPLE} achieves a MAPE of $2.8$--$3.4\times$ lower.

The $k$-fold results confirm the trends observed with LOOCV, with only minor differences due to the different nature of the validation task. 
Together, these results show that \textit{SIMPLE} generalizes well to unseen LLM architectures (LOOCV) and interpolates effectively, accurately predicting unseen workload configurations of known LLMs ($k$-fold).

Finally, on the full dataset, \textit{SIMPLE} reaches an $R^2$ of $0.969$ and a MAPE of $11.974$, confirming the small accuracy gap with \textit{Fine-Grained} observed under cross validation.  The resulting coefficients are used throughout the paper and are reported in Table~\ref{tab:theta} in Appendix~\ref{subsec:appendix-parameters}.

To validate our OPs-centric approach, which aims to give a physical meaning to the fitted coefficients, we also report the coefficients of the OPs-centric \textit{FLOPs-MemOPs} model.
The estimated energy per FLOP ($1.86\times10^{-12}$\,J) and per memory operation ($4.317\times10^{-12}$\,J) are on the order of picoJoules, consistent with the measured energy per operation of Tensor Cores and HBM in NVIDIA GPUs~\cite{gpu_operations_joule}.

\begin{table*}[t]
\centering
\caption{Comparison of $R^2$ and MAPE between \textit{SIMPLE} and LLM-specific baseline models, all fitted with WLS ($w=1/y^2$) on $E_{\text{tok}}$. 
\textit{SIMPLE} is fitted once across all LLMs. As LLM-specific models are fitted separately for each LLM, we report their mean and standard deviation across LLMs.  
}
\label{tab:single-fit}
\smaller

\renewcommand{\arraystretch}{1.5}
\setlength{\tabcolsep}{8pt}

\begin{tabular}{ll>{\hspace{10pt}}l<{\hspace{10pt}}cc}
\toprule
\textbf{Model} & \textbf{$E_{\text{tot}}$ Form} & \textbf{$E_{\text{tok}}$ Formulation} & $\mathbf{R^2}$ & \textbf{MAPE (\%)} \\
\midrule

\makecell[l]{\rule{0pt}{2.6ex}Vellaisamy et al.\ \cite{vellaisamy} \\ Zhan et al.\ \cite{zhan}}
& Linear
& $\theta_0 + \theta_1\dfrac{\tau_{\text{in}}}{\tau_{\text{out}}}$
& \makecell{$0.604$ \\ {\scriptsize $(\pm 0.339)$}}
& \makecell{$38.199$ \\ {\scriptsize $(\pm 10.729)$}}
\\
\midrule

\makecell[l]{\rule{0pt}{2.6ex}Delavande et al.\ \cite{delavande} \\ Wilkins et al.\ \cite{wilkins}}
& Bilinear
& $\theta_0 + \theta_1\dfrac{\tau_{\text{in}}}{\tau_{\text{out}}} + \theta_2\,\tau_{\text{in}}$
& \makecell{$0.886$ \\ {\scriptsize $(\pm 0.065)$}}
& \makecell{$20.477$ \\ {\scriptsize $(\pm 4.890)$}}
\\
\midrule

Oviedo et al.\ \cite{oviedo}
& Power-law
& $\theta_0\,\tau_{\text{in}}^{\theta_1}\,\tau_{\text{out}}^{\theta_2}$
& \makecell{$0.723$ \\ {\scriptsize $(\pm 0.099)$}}
& \makecell{$63.234$ \\ {\scriptsize $(\pm 24.888)$}}
\\
\midrule
\addlinespace[1mm]
SweetSpot \cite{sweetspot}
& Quadratic
& $\theta_{0} + \theta_{1}\dfrac{\tau_{\text{in}}^{2}}{\tau_{\text{out}}} + \theta_{2}\tau_{\text{in}} + \theta_{3}\dfrac{\tau_{\text{in}}}{\tau_{\text{out}}} + \theta_{4}\tau_{\text{out}} + \dfrac{\theta_{5}}{\tau_{\text{out}}}$
& \makecell{$1.000$ \\ {\scriptsize $(\pm 0.000)$}}
& \makecell{$1.841$ \\ {\scriptsize $(\pm 0.692)$}}
\\
\addlinespace[1mm]
\midrule
\addlinespace[2mm]
\textbf{SIMPLE}
& \textbf{Quadratic}

& $\theta_{0} +  
\theta_{1} \, d^2 L \, 
\dfrac{\tau_{\text{in}}}{\tau_{\text{out}}}
+ \theta_{2} \, d \, L \, R
\Big(
    \tau_{\text{in}} + \dfrac{\tau_{\text{out}}}{2}
\Big)
+ \theta_3 d \, L \
$
& $\mathbf{0.969}$
& $\mathbf{11.974}$
\\
\addlinespace[1mm]
\bottomrule
\end{tabular}

\end{table*}

\paragraph{\textbf{Comparison with LLM-specific Models}}
Table~\ref{tab:single-fit} compares the full-dataset fit of \textit{SIMPLE} with SOTA LLM-specific models \cite{vellaisamy, zhan, delavande, wilkins, oviedo, sweetspot}, which are fitted separately for each LLM; for these we report the mean and standard deviation of $R^2$ and MAPE across LLMs in the dataset. 
The previously proposed \textit{SweetSpot}~\cite{sweetspot} model achieves the highest accuracy among SOTA formulations, with a near-perfect in-sample fit (mean $R^2 = 1.000$, mean MAPE $= 1.841\%$).
This sets an upper bound on the accuracy attainable by the proposed LLM-agnostic model. The architecture-aware terms of \textit{SIMPLE} allow it to remain LLM-agnostic: a single set of coefficients covers all architectures, with no per-model fitting. This generality comes at a cost in accuracy, with MAPE increasing from 1.841\% to 11.974\%, while $R^2$ decreases only slightly, from 1.000 to 0.969.

Nevertheless, \textit{SIMPLE} outperforms all other
LLM-specific baselines in both $R^2$ and MAPE. Its MAPE is 41.5\% lower than that of the bilinear models~\cite{delavande,wilkins},
and 3.2$\times$ and 5.3$\times$ lower than those of the
linear~\cite{vellaisamy, zhan} and power-law~\cite{oviedo} models,
resp. The box above summarizes the answer to RQ2.  

\begin{takeaway}{2}{
Can an LLM-agnostic model remain compact and accurate enough to be a practical alternative to LLM-specific models, while generalizing to unseen LLM architectures and workloads?
}
\textit{SIMPLE} shows that a small set of architecture-aware terms captures inference energy without per-LLM fitting. 
Despite using $6.25\times$ fewer coefficients than \textit{Fine-Grained}, it matches its $R^2$ on unseen architectures 
and workloads 
with comparable MAPE.
On the full dataset, it achieves $R^2=0.969$ and MAPE=11.97\%, the best accuracy-compactness trade-off among the evaluated models.
Relative to the most accurate LLM-specific model \textit{SweetSpot}~\cite{sweetspot},
\textit{SIMPLE} trades accuracy for generalization (MAPE 11.97\% vs.\ 1.84\%), while outperforming all other LLM-specific baselines,
making it a practical LLM-agnostic alternative.
\end{takeaway}
\section{Use Cases of the SIMPLE Law}
\label{sec:discussion}
The accuracy and generalizability of \textit{SIMPLE}, demonstrated in Section~\ref{sec:experimental_study}, establish its analytical formulation beyond a predictive model: its architecture-aware terms allow it to be applied as a law, from which structural insights on LLM inference energy can be derived. 
In this section, we apply \textit{SIMPLE} as such to explain a family of notable trade-offs in LLM inference: (i) how LLM size and architecture $f(d, L, R)$  impact $E_{\text{tok}}$ (Section~\ref{sec:sensitivity-analysis}), (ii) 
how input/output sequence $f(\tau_{\text{in}}, \tau_{\text{out}})$ yields an energy minimum in $E_{\text{tok}}$, and how this minimum shapes $E_{\text{tot}}$, and, through it, the Intelligence per Joule attained under increasing reasoning token budgets (Section~\ref{sec:reasoning_workloads}),
 and (iii) how $E_{\text{tot}}$ of a single query shifts from non-agentic to agentic AI workloads (Section~\ref{sec:agentic_workloads}). We use the \textit{SIMPLE} law coefficients fitted on the full dataset (coefficients $\theta$ in Table~\ref{tab:theta}).

To aid readability, we recall here the three contributions of the \textit{SIMPLE} law to $E_{\text{tok}}$, color-highlighted consistently with the figures:

{
\centering
$
\underbrace{%
  \termbox{prefillcol}{\theta_{1}\, d^2 L\, \frac{\tau_{\text{in}}}{\tau_{\text{out}}}}%
}_{\mathclap{\text{\small Prefill Amortization}}}
  \quad
\underbrace{%
  \overbrace{%
    \termbox{kvcol}{\theta_{2}\, d\, L\, R \left( \tau_{\text{in}} + \frac{\tau_{\text{out}}}{2} \right)}%
  }^{\mathclap{\text{\small KV-cache}}}
  \quad
  \overbrace{%
    \termbox{nexttokcol}{\theta_{0} + \theta_{3}\, d\, L}%
  }^{\mathclap{\text{\small Next-token}}}
}_{\mathclap{\text{\small Decode Accumulation}}}
$
\par
}

\begin{figure*}[t]
  \centering
  \includegraphics[width=\textwidth]{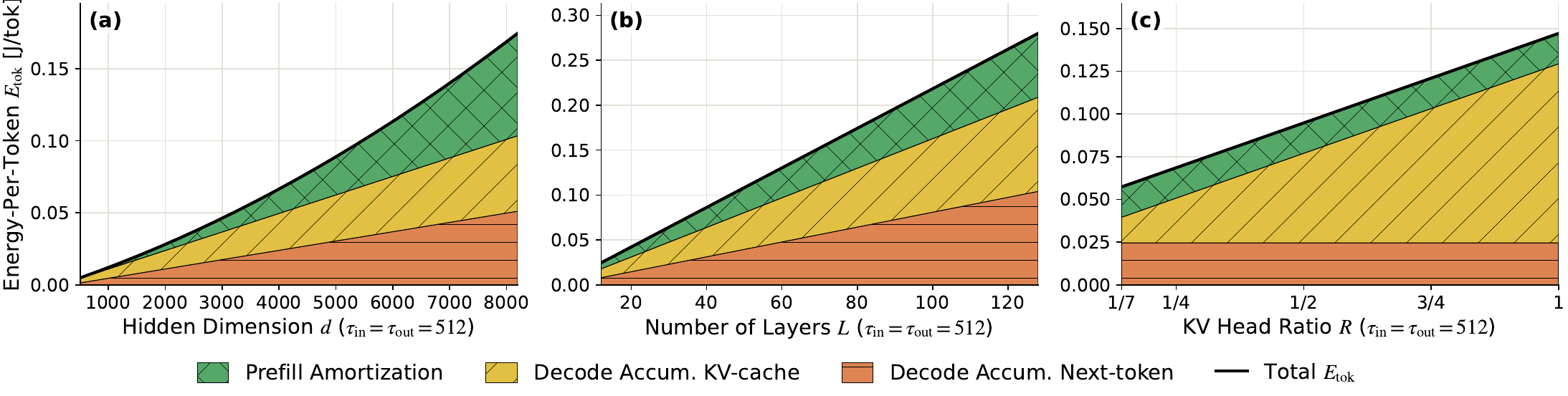}
  \caption{Term-wise decomposition of the predicted $E_{\text{tok}}$ under the \textit{SIMPLE} law, varying hidden dimension $d$ (a), number of layers $L$ (b), and KV head ratio $R$ (c), using Llama~3.1~8B architectural hyperparameters, with $\tau_{\text{in}} = \tau_{\text{out}} = 512$. Each band represents the additive contribution of one term of the \textit{SIMPLE} law.}
  \label{fig:lineplot}
\end{figure*}
\begin{figure}
\centering
\includegraphics[width=\columnwidth]{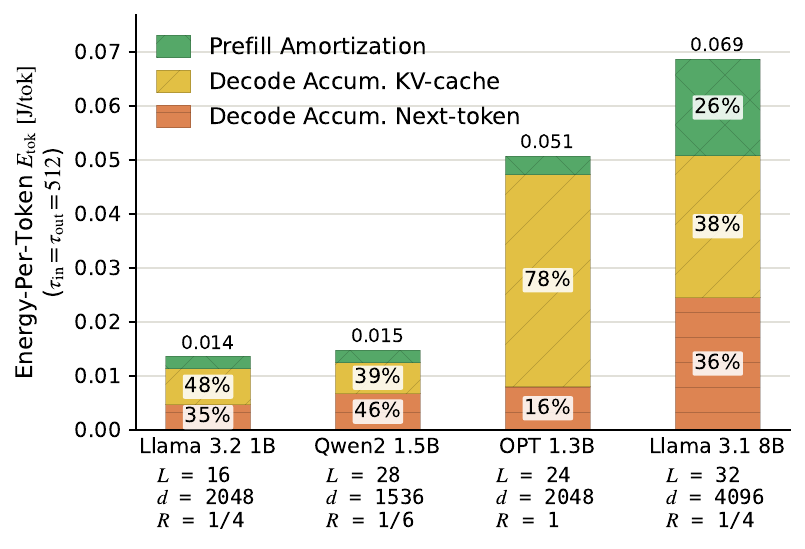}
    \caption{
    Term-wise decomposition of the predicted $E_{\text{tok}}$ under the \textit{SIMPLE} law, using Llama~3.2~1B, Qwen2~1.5B,  OPT~1.3B, and 
    Llama~3.1~8B,  with $\tau_{\text{in}} = \tau_{\text{out}} = 512$.
    }
    \label{fig:share_models}  
\end{figure}

\subsection{Impact of LLM Size and Architecture on Energy}
\label{sec:sensitivity-analysis}
Figure~\ref{fig:lineplot} shows the contribution of each \textit{SIMPLE} law term 
to the predicted $E_{\text{tok}}$ as a function of the individual key architectural hyperparameters: the hidden dimension $d$, the number of layers $L$, and the KV-to-query head ratio $R = n_{\text{kv}}/n_{\text{q}}$. We set $\tau_{\text{in}} = \tau_{\text{out}} = 512$ and use Llama~3.1 8B architectural hyperparameters as a reference.
 
In Figure~\ref{fig:lineplot}~(a), the Prefill Amortization term 
grows quadratically with $d$, while the Decode Accumulation terms grow only linearly: the KV-cache term is proportional to $d$, and the Next-token term is affine in $d$ due to the constant offset $\theta_0$.  This quadratic dependence dominates the overall scaling, producing a rapid increase in $E_{\text{tok}}$ for large hidden dimensions and marking $d$ as one of the most sensitive LLM architectural design hyperparameter. Moreover, it suggests that the energy penalty of LLMs with a larger $d$ is amplified in prefill-dominated scenarios where long contexts are paired with short outputs.

The number of layers $L$, shown in Figure~\ref{fig:lineplot}~(b),
multiplies the Prefill Amortization and KV-cache terms, and the $\theta_3 d L$ component of the Next-token term. Only the constant offset $\theta_0$ is independent of $L$, so $E_{\text{tok}}$ grows proportionally with model depth.
Finally, Figure~\ref{fig:lineplot}~(c) shows the energy impact of the attention head configuration. The Decode Accumulation KV-cache 
is the only term that depends on $R$. It scales linearly with $\tau_{\text{in}}$ and $\tau_{\text{out}}$, so 
a smaller $R$ (i.e., a larger $1/R = n_{\text{q}}/n_{\text{kv}}$, corresponding to more aggressive GQA grouping) reduces the KV-cache cost during  decoding, making LLM architectures with higher GQA grouping more efficient on long output sequences. In the given example, $E_{\text{tok}}$ is $\approx$60~mJ/tok under 
$R=1/7$, compared to $\approx$150~mJ/tok
under standard MHA ($R=1$), a reduction of approximately $3\times$.
How do these dependencies impact the token energy consumption across (i) LLMs of the same family with growing size and (ii) LLMs of the same size but different families?
\begin{takeaway}{3}{How do LLM size and architectural hyperparameters influence inference energy consumption?}
\textit{SIMPLE} law shows that hidden dimension $d$ is the most sensitive architectural hyperparameter, with prefill energy scaling quadratically with $d$, while layer depth $L$ contributes linearly. Attention architecture matters strongly: more aggressive GQA KV grouping ($1/R=7$) reduces $E_{\text{tok}}$ by up to $3\times$ relative to standard MHA ($R=1$), identifying it as an impactful architectural choice for reducing inference energy. 
Applied across LLMs of different sizes and families, it explains how the energy cost changes and which term accounts for it.
These results confirm that architectural hyperparameters are key determinants
of inference energy efficiency, capturing effects that model size alone cannot explain.
\end{takeaway}

Figure~\ref{fig:share_models} shows the contribution of each \textit{SIMPLE} term to the predicted $E_{\text{tok}}$ at $\tau_{\text{in}} = \tau_{\text{out}} = 512$, using Llama~3.2~1B  as the reference and comparing it against three other models: Qwen2~1.5B and OPT~1.3B, which are similar in size, and Llama~3.1~8B, which belongs to the same family but is larger. Qwen2~1.5B shows an $E_{\text{tok}}$ similar to Llama~3.2~1B (15 vs.\ 14~mJ/tok) but a different breakdown: its smaller $d$ (1536 vs. 2048) and $R$ (1/6 vs. 1/4), partially offset by a larger $L$ (28 vs. 16), reduce the KV-cache share from 48\% to 39\%; the combined effect of its smaller $d$ and larger $L$ keeps the Prefill Amortization nearly unchanged, while increasing the Next-token share from 35\% to 46\%. At longer sequences, these structural differences favor Qwen2~1.5B, whose lower KV-cache coefficient increasingly dominates, making it more energy-efficient than Llama~3.2~1B.
A starker contrast is seen with OPT~1.3B, which has the same $d$ as Llama~3.2~1B but a larger $L$ (24 vs. 16) and uses MHA ($R=1$) instead of GQA ($R=1/4$).  As a result, the KV-cache share grows from 48\% to 78\%, increasing the KV-cache energy from 6~mJ/tok to 40~mJ/tok, resulting in an overall $E_{\text{tok}}$ increase from 14~mJ/tok to 51~mJ/tok. Comparing Llama~3.2~1B with the larger Llama~3.1~8B, the Prefill Amortization share increases from 17\% to 26\% while the KV-cache share decreases from 48\% to 38\%: as both $d$ (4096 vs. 2048) and $L$ (32 vs. 16) double, Prefill Amortization grows $8 \times$ while the KV-cache term grows only $4\times$, due to the quadratic contribution of $d$, despite the same $R$. 

These results show that the \textit{SIMPLE} law explains the energy differences between LLMs (Figure~\ref{fig:average_efficiency}) as well as their expected scalability to longer input/output sequences through the architectural hyperparameters that drive each inference phase, which model size alone cannot capture. The box above summarizes the answer to RQ3. 
\begin{figure*}[th]
    \centering
    \subfloat[Qwen3-1.7B\\Architectural hyperparameters: $d=2048$, $L=28$, $R=8/16$. \label{fig:E_tok_contour_n_in_n_out_qwen3_1.7b}]{
        \includegraphics[width=0.48\textwidth]{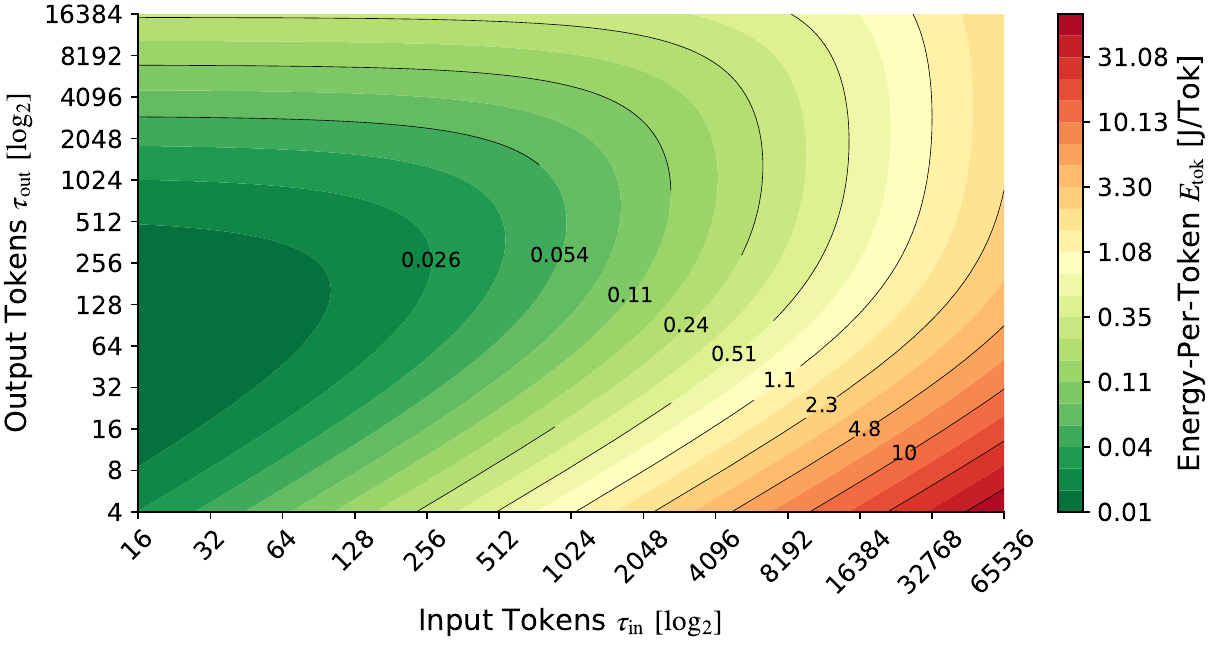}
    }\hfill
    \subfloat[DeepSeek-R1-Distill-Llama-8B\\Architectural hyperparameters: $d=4096$, $L=32$, $R=8/32$. \label{fig:E_tok_contour_n_in_n_out_ds_llama_8b}]{
        \includegraphics[width=0.48\textwidth]{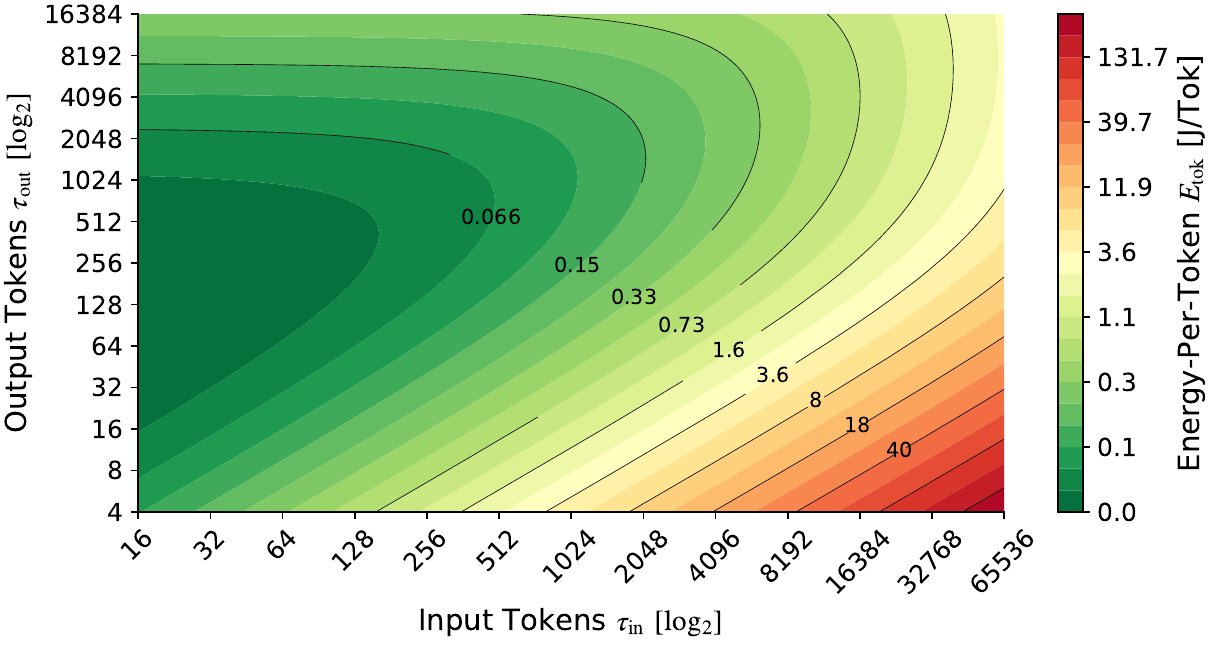}}
    \caption{Estimated $E_{\text{tok}}$ contour maps generated by the \textit{SIMPLE} law of Qwen3-1.7B (a) and DeepSeek-R1-Distill-Llama-8B (b).}
    \label{fig:sweetspot_contours}
\end{figure*}
\begin{figure}
\centering
\includegraphics[width=\columnwidth]{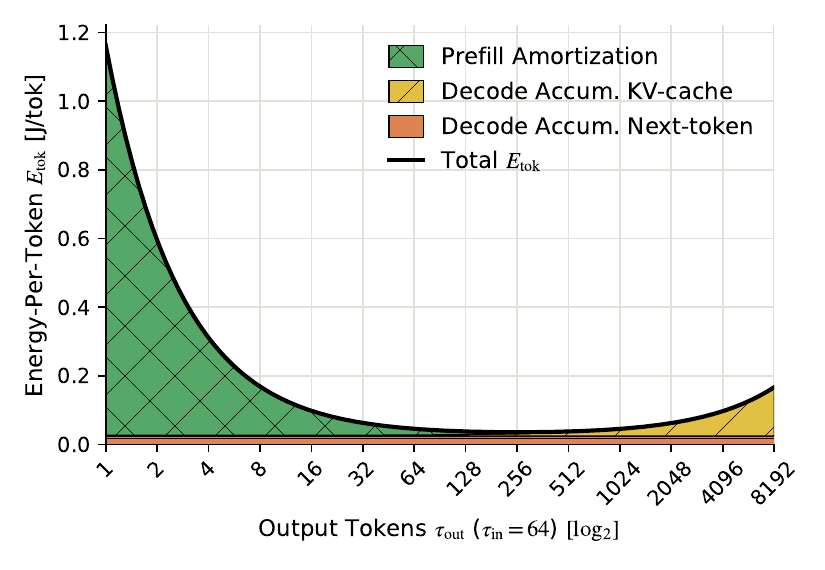}
    \caption{
    $E_{\text{tok}}$ additive components of DeepSeek-R1-Distill-Llama-8B predicted by the \textit{SIMPLE} law, with $\tau_{\text{in}} = 64$.
    }
    \label{fig:sweetspot}  
\end{figure}
\subsection{From Energy Minimum to Reasoning Efficiency}
\label{sec:reasoning_workloads}
Throughout, 
Qwen3-1.7B and \seqsplit{DeepSeek-R1-Distill-Llama-8B} to derive the $E_{\text{tok}}$ minimum analytically and to map $E_{\text{tok}}$ over a wide input/output token range and connect $E_{\text{tot}}$ to Intelligence per Joule. 

\paragraph{\textbf{$E_{\text{tok}}$ and Input/Output Sequence}}

Figure~\ref{fig:sweetspot} shows the predicted  $E_{\text{tok}}$ for DeepSeek-R1-Distill-Llama-8B as a function of $\tau_{\text{out}}$, with $\tau_{\text{in}}$ fixed at an arbitrary length of 64. The colored regions correspond to the additive components of the \textit{SIMPLE} law (Prefill Amortization, and Decode Accumulation, comprising separate KV-cache and Next-token regions), with the black curve giving the total.
The Prefill Amortization energy cost 
decreases as $1/\tau_{\text{out}}$, so its $E_{\text{tok}}$  decreases hyperbolically as more output tokens are generated. In contrast, the Decode Accumulation cost of the KV-cache term
increases linearly with $\tau_{\text{out}}$, reflecting the accumulation of KV-cache entries with each new output token. The remaining cost of the Next-token term 
is independent of the output length and contributes a constant energy. The balance between the decreasing prefill contribution and the increasing decode contribution produces a minimum in the $E_{\text{tok}}$ curve, corresponding to the \emph{sweet spot}~\cite{sweetspot} of inference energy efficiency. 
The location of this optimum can be obtained analytically from the \textit{SIMPLE} law by differentiating $E_{\text{tok}}$ with respect to $\tau_{\text{out}}$ and setting the derivative to zero. Since $d$ and $L$ are strictly positive by definition, the common factor $dL>0$ can be divided out, yielding:

{
\small
\[
\frac{\partial E_{\text{tok}}}{\partial \tau_{\text{out}}} 
= 
-\,\theta_1 d\,\frac{\tau_{\text{in}}}{\tau_{\text{out}}^2} 
+ \theta_2 \, \frac{R}{2} = 0
\]
}
\noindent
Solving for \(\tau_{\text{out}}\) yields the optimal output sequence length:

{
\small
\begin{equation}
\label{eq:sweet-spots}
\tau_{\text{out}}^\ast = \sqrt{2 \, \frac{\theta_1}{\theta_2} \, \frac{1}{R}\,d\,\tau_{\text{in}}} = 
\sqrt{2\, \frac{\theta_1}{\theta_2} \, \frac{n_{\text{q}}}{n_{\text{kv}}} \, d\,\tau_{\text{in}}}
\end{equation}
}
\noindent
This result gives a closed-form expression for the \emph{sweet spot}, i.e., the output sequence length that minimizes the energy per generated token. In particular, the optimal output length grows proportionally to $\sqrt{\tau_{\text{in}}d}$, and is inversely proportional to the KV-to-query head ratio $\sqrt{R} = \sqrt{n_{\text{kv}}/n_{\text{q}}}$.  Consequently, larger input lengths require longer generations to amortize the prefill cost, while architectures with a larger $R$ reach the energy optimum with shorter outputs.

Figure~\ref{fig:sweetspot_contours} shows the $E_{\text{tok}}$ landscape generated by the \textit{SIMPLE} law, over the input  token range $[16,65536]$ and output token range $[4,16384]$, using Qwen3-1.7B and DeepSeek-R1-Distill-Llama-8B architectural hyperparameters. These spans cover the LLM workloads used in conversation and coding \cite{dynamo}, agentic \cite{liu2026agentic} and reasoning \cite{s1-test-time, reasoning_swi}.
Both predicted landscapes closely mirror the non-linear efficiency pattern empirically observed  in the dataset of \cite{sweetspot} (Figure~\ref{fig:heatmap_average}), capturing both the prefill-amortization and the decode-dominated regimes within a single surface. For a fixed $\tau_{\text{in}}$, the surface exhibits the expected amortization effect, with $E_{\text{tok}}$ 
reaching a minimum before rising again as the linear decode cost dominates.  
For the shortest input ($\tau_{\text{in}}=16$), this minimum lies at $\tau_{\text{out}}\approx64$ tokens for Qwen3-1.7B and $\tau_{\text{out}}\approx128$ for DeepSeek-R1-Distill-Llama-8B, and corresponds to the most energy-efficient point in the token space, at $\approx$12 and $\approx$30~mJ/tok, respectively.  As $\tau_{\text{in}}$ increases, the contours elongate along the output axis, reflecting the growing weight of the prefill term relative to the linear decoding contribution.  The effect is strongest for short outputs, where the prefill cost is amortized over only a handful of tokens: the least energy-efficient point in the token space is located at $\tau_{\text{in}}=65\text{K}$ and $\tau_{\text{out}}=4$, where $E_{\text{tok}}$ reaches $\approx$66~J/tok for Qwen3-1.7B and $\approx$293~J/tok for DeepSeek-R1-Distill-Llama-8B. 

\paragraph{\textbf{Reasoning Cost and Intelligence per Joule}} 
\begin{table}[t]
\centering
\footnotesize
\caption{Accuracy on AIME-24 dataset from \cite{reasoning_swi} and \textit{SIMPLE} estimation of $E_{\mathrm{tok}}$ and $E_{\mathrm{tot}}$ for Qwen3-1.7B and DeepSeek-R1-Distill-Llama-8B as a function of reasoning token budgets. Ratios (in parentheses) are relative to the minimum budget, corresponding to $\tau_{\mathrm{out}}=2048$. Intelligence per Joule (IPJ) is computed as accuracy (\%) divided by $E_{\mathrm{tot}}$ (J). For both models, the mean $\tau_{\mathrm{in}}$ of the dataset is used ($\tau_{\text{in}}=105$).}
\label{tab:acc_energy}
\renewcommand{\arraystretch}{1.15}
\setlength{\tabcolsep}{5.2pt}
\begin{tabular}{rrrrr}
\toprule
 \multicolumn{1}{c}{\textbf{Reasoning}}
 & \multicolumn{1}{c}{\textbf{Acc.}}
 & \multicolumn{1}{c}{$\boldsymbol{E_{\mathrm{tok}}}$}
 & \multicolumn{1}{c}{$\boldsymbol{E_{\mathrm{tot}}}$}
 & \multicolumn{1}{c}{\textbf{IPJ}} \\
 \multicolumn{1}{c}{[tok]}
 & \multicolumn{1}{c}{\scriptsize [\%]}
 & \multicolumn{1}{c}{\scriptsize [mJ/tok]}
 & \multicolumn{1}{c}{\scriptsize [J]}
 & \multicolumn{1}{c}{\scriptsize [\%/kJ]} \\
\midrule
\multicolumn{5}{@{}l}{\textbf{Qwen3-1.7B}} \\
\addlinespace[3pt]
 2048  {\tiny ($1.0\times$)} &  3.0 {\tiny ($1.00\times$)}  &  43.6 {\tiny ($1.00\times$)} &   89.3 {\tiny ($1.00\times$)}  & 33.7 {\tiny ($1.00\times$)} \\
 4063  {\tiny ($2.0\times$)} & 14.0 {\tiny ($4.67\times$)}  &  73.6 {\tiny ($1.69\times$)} &  298.9 {\tiny ($3.35\times$)}  & 46.8 {\tiny ($1.39\times$)} \\
 6978  {\tiny ($3.4\times$)} & 27.0 {\tiny ($9.00\times$)}  & 117.0 {\tiny ($2.68\times$)} &  816.5 {\tiny ($9.14\times$)}  & 33.1 {\tiny ($0.98\times$)} \\
 9510  {\tiny ($4.6\times$)} & 38.0 {\tiny ($12.67\times$)} & 154.8 {\tiny ($3.55\times$)} & 1471.9 {\tiny ($16.48\times$)} & 25.8 {\tiny ($0.77\times$)} \\
 11896 {\tiny ($5.8\times$)} & 46.0 {\tiny ($15.33\times$)} & 190.4 {\tiny ($4.37\times$)} & 2264.5 {\tiny ($25.36\times$)} & 20.3 {\tiny ($0.60\times$)} \\
\midrule
\multicolumn{5}{@{}l}{\textbf{DeepSeek-R1-Distill-Llama-8B}} \\
\addlinespace[3pt]
 2048  {\tiny ($1.0\times$)} &  2.0 {\tiny ($1.00\times$)}  &  64.0 {\tiny ($1.00\times$)} &  131.1 {\tiny ($1.00\times$)}  & 15.3 {\tiny ($1.00\times$)} \\
 3515  {\tiny ($1.7\times$)} & 15.0 {\tiny ($7.50\times$)}  &  88.7 {\tiny ($1.39\times$)} &  311.6 {\tiny ($2.38\times$)}  & 48.1 {\tiny ($3.15\times$)} \\
 5086  {\tiny ($2.5\times$)} & 28.0 {\tiny ($14.00\times$)} & 115.3 {\tiny ($1.80\times$)} &  586.3 {\tiny ($4.47\times$)}  & 47.8 {\tiny ($3.13\times$)} \\
 7330  {\tiny ($3.6\times$)} & 41.0 {\tiny ($20.50\times$)} & 153.4 {\tiny ($2.40\times$)} & 1124.6 {\tiny ($8.58\times$)}  & 36.4 {\tiny ($2.39\times$)} \\
 8145  {\tiny ($4.0\times$)} & 44.0 {\tiny ($22.00\times$)} & 167.3 {\tiny ($2.61\times$)} & 1362.6 {\tiny ($10.39\times$)} & 32.3 {\tiny ($2.11\times$)} \\
\bottomrule
\end{tabular}
\end{table}

\begin{figure}
\centering\includegraphics[width=\columnwidth]{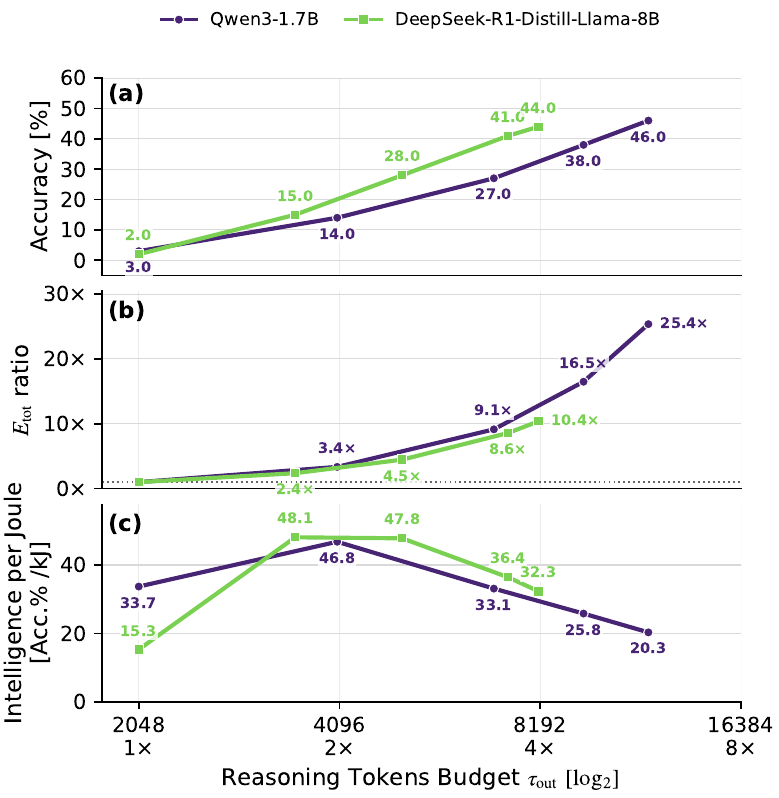}
    \caption{
    Graphical comparison of the accuracy (a), $E_\text{tot}$ (b) and Intelligence per Joule (c) results reported in Table~\ref{tab:acc_energy}.
    }
    \label{fig:acc_E_tok}  
\end{figure}
Reasoning (or thinking) is widely referred to as the process of generating intermediate tokens before producing the final response, thereby leveraging additional test-time computation to improve accuracy \cite{s1-test-time, reasoning_swi}. 
Throughout this section, we refer to the total number of generated tokens, including explicit reasoning tokens, as the   \textit{reasoning token budget}. Since \textit{SIMPLE} already models all generated tokens as $\tau_\text{out}$, increasing the reasoning token budget directly increases $\tau_\text{out}$ and hence $E_{\text{tot}}$. To assess how this energy cost translates into reasoning efficiency, we combine the $E_{\text{tot}}$ predicted by \textit{SIMPLE} with task accuracy, measured as Intelligence per Joule (IPJ~$=$~accuracy~$/~E_{\text{tot}})$. We draw on results from \cite{reasoning_swi}, which reported the accuracy of a set of SOTA LLMs (including Qwen3-1.7B \cite{qwen3} and DeepSeek-R1-Distill-Llama-8B \cite{deepseek-r1}) on widely used benchmarks while varying reasoning budget.

From the results of \cite{reasoning_swi}, we report in Table~\ref{tab:acc_energy} the accuracy as a function of the reasoning token budget
for Qwen3-1.7B and DeepSeek-R1-Distill-Llama-8B on the AIME-24 dataset~\cite{aime24}, for which both models report results across a comparable range of reasoning token budgets. In the same table, we also report the estimated $E_\text{tok}$ and $E_\text{tot}$ using the \textit{SIMPLE} law, at the corresponding reasoning token budget, together with their ratios with respect to the minimum considered reasoning token budget, 
$\tau_\text{out}=2048$. 
As input length, we use AIME-24 mean prompt length ($\tau_{\text{in}}=105$). 

Figure~\ref{fig:acc_E_tok} shows that 
accuracy increases consistently with the reasoning token budget $\tau_\text{out}$ for both models, as does $E_\text{tot}$.
For Qwen3-1.7B, accuracy rises from 3.0\% to 46.0\% as $\tau_\text{out}$ grows from 2048 up to 11896 tokens, corresponding to a reasoning budget $5.8\times$ the minimum. Over the same range, $E_\text{tok}$ rises from 43.6~mJ/tok up to 190.4~mJ/tok, increasing $E_\text{tot}$ by $25.4\times$ and lowering IPJ to $0.60\times$ of its initial value. DeepSeek-R1-Distill-Llama-8B shows a similar trend: accuracy improves from 2.0\% to 44.0\% as $\tau_\text{out}$ grows from 2048 to 8145 tokens, increasing $E_\text{tot}$ by $10.4\times$. Unlike Qwen3-1.7B, this causes an IPJ increase of $2.11\times$.  

Despite these opposite trends at the two reasoning extremes, both models show an IPJ peak neither at the minimum nor the maximum reasoning budget, but at an intermediate level, as visible in Figure~\ref{fig:acc_E_tok}(c). This arises because all reported budgets lie well past the $E_\text{tok}$ minimum predicted by \textit{SIMPLE} ($\tau_\text{out}^* \approx 165$ tokens for Qwen3-1.7B and $\approx 331$ tokens for DeepSeek-R1-Distill-Llama-8B), so $E_\text{tot}$ grows super-linearly as $E_\text{tok}$ and $\tau_\text{out}$ increase together, while the marginal accuracy gain per additional token decreases. 
Concretely, at the IPJ peak, the accuracy gain exceeds the $E_\text{tot}$ increase for both models; for Qwen3-1.7B (4063 tokens), accuracy grows by 4.67$\times$ against $E_\text{tot}$ by 3.35$\times$, yielding an IPJ gain of 1.39$\times$; for DeepSeek-R1-Distill-Llama-8B (3515 tokens), 7.5$\times$ against 2.38$\times$, yielding an IPJ gain of 3.15$\times$.
\begin{takeaway}{4}{How does the reasoning token budget affect inference energy and Intelligence per Joule?}
The \textit{SIMPLE} law analytically derives the output length $\tau_{\text{out}}^*$ that minimizes per-token energy (the \textit{sweet spot}) showing that practical reasoning token budgets lie well beyond this minimum, driving super-linear growth
in $E_{\text{tot}}$. Although accuracy grows with the reasoning budget, energy grows faster: Intelligence per Joule peaks at an intermediate budget and falls by up to $3.14\times$
at the lowest and highest budgets across a $4.0\times$ range.
\end{takeaway}

This application of the \textit{SIMPLE} law provides a quantitative framework for identifying the reasoning budget that maximizes IPJ. The box above summarizes the RQ4 answer.  

\subsection{Energy Cost in Agentic Workloads }
\label{sec:agentic_workloads}

So far, SOTA works \cite{delavande, oviedo, google} have focused on evaluating the environmental impact of conversational AI. Agentic AI workloads are changing this landscape, and no existing work proposes a theoretical framework that can capture and explain the resulting transformation in energy cost, as they lack a way of accounting for prompt/prefix caching.
Liu et al. \cite{liu2026agentic} present an extensive analysis of the GitHub Copilot \cite{copilot2025github} coding agent production workload. Unlike conversational AI, characterized by queries with a median prompt length ($\tau_{\text{in}}$) of 750--1050 tokens and a median completion length ($\tau_{\text{out}}$) of 80--105, the reported agentic workload follows a long-tail distribution with a median context length ($\tau_{\text{in}}$) of 68K tokens and a median completion length of 247 tokens. Moreover, they report a median prefix KV-cache of 63K tokens, which corresponds to a hit rate of $\approx93\%$.
This means that in the average query only 5K input tokens (out of 68K) are newly computed in the prefill stage. 

We apply the \textit{SIMPLE} law with the following adaptation. 
Let $\tau_{\text{c}} \le \tau_{\text{in}}$ be the number of input tokens served from the cache. Only the uncached suffix needs to be prefilled, so the Prefill Amortization term 
scales with $\tau_{\text{in}}-\tau_{\text{c}}$ instead of $\tau_{\text{in}}$.

Table~\ref{tab:agentic_prefix_cache_etot} quantifies this effect for Llama~3.1~8B on the median GitHub Copilot request, comparing three scenarios. In the non-agentic scenario, for which we consider only the non-cached input ($\tau_{\text{in}}$= $5$K, $\tau_{\text{c}}$= $0$), the request costs 138.1~J, with Prefill (88.8~J) as the dominant term and the KV-cache contribution (43.2~J) playing a secondary role. Moving to the agentic scenario ($\tau_{\text{in}}$= 68K, $\tau_{\text{c}}$= 0), both context-dependent terms grow approximately proportionally to $\tau_{\text{in}}$: Prefill increases by $13.6\times$ and KV-cache by $13.3\times$, while the Next-token term is unchanged, bringing $E_{\text{tot}}$ to 1787.7~J ($13.0\times$). Prefix caching ($\tau_{\text{in}}$= 68K, $\tau_{\text{c}}$= 63K) is highly effective on the Prefill term in this setting: only the 5K uncached tokens are processed, reducing Prefill contribution back to 88.8~J ($-93\%$) and reducing $E_{\text{tot}}$ by $2.7\times$ (668.7~J). However, the agentic cached request still consumes $4.8\times$ more energy than the non-agentic one. This is because the KV-cache term is unaffected by prefix caching: the full 68K context must still be read and attended to at every decoding step, accounting for 86\% of the remaining energy.
\begin{table}[t]
\centering
\footnotesize
\caption{\textit{SIMPLE} law estimation of the per-request energy, decomposed by contributions, for Llama-3.1-8B on the median GitHub Copilot coding-agent request from \cite{liu2026agentic} ($\tau_{\mathrm{out}}=247$). In parentheses, the ratio w.r.t.\ the non-agentic case. $\Delta E_{\mathrm{tot}}$ is the energy cost of one additional token ($\tau_{\mathrm{out}}=248$).}
\label{tab:agentic_prefix_cache_etot}
\renewcommand{\arraystretch}{1.15}
\setlength{\tabcolsep}{4pt}
\begin{tabular}{@{}lrrr@{}}
\toprule
\shortstack[l]{\textbf{Term}\\{\scriptsize [J]}}
 & \shortstack[r]{\textbf{Non-Agentic}\\{\scriptsize $\tau_{\mathrm{in}}{=}5$K, $\tau_{\mathrm{c}}{=}0$}}, 
 & \shortstack[r]{\textbf{Agentic}\\{\scriptsize $\tau_{\mathrm{in}}{=}68$K, $\tau_{\mathrm{c}}{=}0$}}
 & \shortstack[r]{\textbf{Agentic Cached}\\{\scriptsize $\tau_{\mathrm{in}}{=}68$K, $\tau_{\mathrm{c}}{=}63$K}} \\
\midrule
Prefill    &  88.8 {\scriptsize ($1.0\times$)} & 1207.8 {\scriptsize ($13.6\times$)} &  88.8 {\scriptsize ($1.0\times$)} \\
KV-cache   &  43.2 {\scriptsize ($1.0\times$)} &  573.8 {\scriptsize ($13.3\times$)} & 573.8 {\scriptsize ($13.3\times$)} \\
Next-token &   6.1 {\scriptsize ($1.0\times$)} &    6.1 {\scriptsize ($1.0\times$)} &   6.1 {\scriptsize ($1.0\times$)} \\
$E_{\mathrm{tok}}$ {\scriptsize [J/tok]} & 0.56 {\scriptsize ($1.0\times$)} & 7.24 {\scriptsize ($12.9\times$)} & 2.71 {\scriptsize ($4.8\times$)} \\
\textbf{Total} $\boldsymbol{E_{\mathrm{tot}}}$ & \textbf{138.1} {\scriptsize ($\mathbf{1.0\times}$)} & \textbf{1787.7} {\scriptsize ($\mathbf{13.0\times}$)} & \textbf{668.7} {\scriptsize ($\mathbf{4.8\times}$)} \\
\midrule
$\boldsymbol{\Delta E_{\mathrm{tot}}}$ (+1 tok) & \textbf{0.20} {\scriptsize ($\mathbf{1.0\times}$)} & \textbf{2.35} {\scriptsize ($\mathbf{11.8\times}$)} & \textbf{2.35} {\scriptsize ($\mathbf{11.8\times}$)} \\
\bottomrule
\end{tabular}
\end{table}

\begin{takeaway}{5}{How does the energy cost of a single query shift from non-agentic to agentic workloads?}
Moving from a non-agentic ($\tau_{\text{in}}=5$K) to an agentic ($\tau_{\text{in}}=68$K) request raises the energy per request by $13.0\times$, as both the prefill and KV-cache terms grow with $\tau_{\text{in}}$. Prefix caching removes most of the prefill cost ($-93\%$, $2.7\times$ lower $E_{\text{tot}}$), but the KV-cache term, which accounts for 86\% of the remaining energy, is unaffected: a cached agentic request still costs $4.8\times$ more than a non-agentic one. Likewise, each additional generated token costs $11.8\times$ more, with or without caching.
\end{takeaway}

The last row reports the energy cost of generating one additional token ($\tau_{\text{out}}$= 248). For the non-agentic request, it is 0.20~J. In the agentic scenario, it rises to 2.35~J ($11.8\times$), driven by KV-cache and Next-token energy growth, while the Prefill energy is unchanged. Since caching only affects Prefill, the cost of the additional generated token is identical in the agentic scenario with or without prefix caching. The box above summarizes the RQ5 answer.
\section{Conclusions}
\label{sec:conclusions}

This paper presented the \textit{SIMPLE} law, an architecture-aware closed-form expression for predicting the energy cost of dense decoder-only LLM inference, derived from the computational and memory-access complexity of the prefill and decode phases. 
\textit{SIMPLE} expresses energy consumption with a compact and interpretable formulation, which explains prefill and decode energy costs as a function of LLM architectural hyperparameters and input/output sequence lengths. By explicitly modeling the attention mechanism, \textit{SIMPLE} closes a gap in the literature, as the first analytical energy law able to generalize across LLM architectures.  With only four parameters fitted once across all LLMs, it reduces the error of previous SOTA models by 3.1--3.5$\times$, reaching a MAPE of 11.97\%.

Beyond predictive accuracy, the \textit{SIMPLE} law quantitatively explains three of the most prominent effects in modern LLM inference: how LLM size and architecture affect energy cost, 
how the reasoning token budget drives super-linear energy growth and shapes the Intelligence per Joule,  and how agentic AI transforms energy consumption. Taken together, these results establish the \textit{SIMPLE} law as an analytical foundation for sustainable LLM inference.

Our approach is subject to a few limitations, which we group into three categories.  First, our evaluation is restricted to dense, single-GPU LLMs and does not cover mixture-of-experts models.  
We also restrict our evaluation to MHA and GQA attention mechanisms; validation on more recent attention variants, LLM sizes, and architectures is left for future work. Second, the \textit{SIMPLE} law does not explicitly model batch size or the number of concurrent requests as hyperparameters of the formulation. 
We plan to explicitly incorporate such additional workload parameters to extend the law. Finally, the reference dataset was collected using TensorRT-LLM on a single H100 GPU. Therefore, the \textit{SIMPLE} law captures the energy behavior of single-device inference, and does not account for the additional costs of multi-GPU deployments, such as inter-device communication, tensor or pipeline parallelism, and distributed KV-cache management. As such, the fitted coefficients are specific to this hardware-framework pair. How the coefficients would change using other accelerators, multi-GPU deployments, and other inference frameworks remains to be explored in future work.

\section*{Acknowledgements}
The research has been partially funded by EuroHPC JU SEANERGYS (g.a. 101177590), DARE (g.a. 101143421) and Spoke “FutureHPC \& BigData” of the ICSC.
\appendix
\section{Coefficients and Predictions}
\label{subsec:appendix-parameters}
Table~\ref{tab:theta} reports the values of \textit{SIMPLE} and the OPs-centric \textit{FLOPs-MemOPs} models coefficients $\theta$, fitted over the full dataset, together with their standard errors, $t$-statistics, and $p$-values, obtained using WLS ($w=1/y^2$).

\begin{table}[!b]
\small
\centering
\caption{Fitted coefficients of the \textit{SIMPLE} Law ($R^2=0.969$, MAPE $=11.974\%$) and of the OPs-centric \textit{FLOPs-MemOPs} model ($R^2=0.581$, MAPE $=39.384\%$).}
\label{tab:theta}
\renewcommand{\arraystretch}{1.1}
\begin{tabular}{@{}lrrrr@{}}
\toprule
Coeff. & Estimate & Std.\ Err. & $t$ & $P{>}|t|$\\
\midrule
\multicolumn{5}{@{}l}{\textbf{\textit{SIMPLE}}} \\
\addlinespace[2pt]
$\theta_0$ & $-1.866\times10^{-3}$ & $0.000$              & $-4.412$ & $<0.001$ \\
$\theta_1$ & $3.308\times10^{-11}$ & $6.80\times10^{-13}$ & $48.631$ & $<0.001$ \\
$\theta_2$ & $1.041\times10^{-9}$  & $1.24\times10^{-11}$ & $84.004$ & $<0.001$ \\
$\theta_3$ & $2.020\times10^{-7}$  & $8.01\times10^{-9}$  & $25.208$ & $<0.001$ \\
\midrule
\multicolumn{5}{@{}l}{\textbf{\textit{FLOPs-MemOPs}}} \\
\addlinespace[2pt]
$\theta_0$ & $7.057\times10^{-3}$  & $9.33\times10^{-4}$  & $7.566$  & $<0.001$ \\
$\theta_1$ & $1.856\times10^{-12}$ & $9.19\times10^{-14}$ & $20.196$ & $<0.001$ \\
$\theta_2$ & $4.317\times10^{-12}$ & $5.98\times10^{-13}$ & $7.221$  & $<0.001$ \\
\bottomrule
\end{tabular}
\end{table}

\bibliographystyle{unsrtnat}
\bibliography{ref}

\end{document}